\documentclass[runningheads]{llncs}
\usepackage{geometry}
\usepackage[font = small, margin=2cm]{caption}

\usepackage{type1cm}        
\usepackage{makeidx}         
\usepackage{graphicx}        
\usepackage{multicol}        
\usepackage[bottom]{footmisc}

\usepackage{soul}
\usepackage{newtxtext}       %
\usepackage{newtxmath}       

\usepackage{float}
\usepackage{xurl}

\usepackage{subfigure}
\usepackage[ruled,vlined]{algorithm2e}
\usepackage{algorithmicx}
\usepackage{mathrsfs}
\usepackage{algpseudocode}
\usepackage{cprotect}

\usepackage{color}
\definecolor{Green}{rgb}{.1,.75,.55}

\definecolor{Purple}{rgb}{.55,.0,.55}

\definecolor{Blue}{rgb}{.0,.0,.80}

\usepackage{parskip} 
\makeindex             

\begin{document}

		\title{\emph{\small{*Preprint version (before any peer-review) of a book chapter to appear in the Book series -- Studies in Computational Intelligence, Book title -- Federated Learning Systems: Towards Next-generation AI, Book eds. Muhammad Habib ur Rehman and {Mohamed~Medhat~Gaber}, Publisher -- Springer Nature Switzerland AG
					Gewerbestrasse 11, 6330 Cham, Switzerland.}} \\
			Advancements of federated learning towards privacy preservation: from federated learning to split learning}
		
		\author{Chandra Thapa \and M.A.P. Chamikara \and Seyit~A. Camtepe}
		\institute{CSIRO Data61, Australia \\
			\email{\{chandra.thapa, chamikara.arachchige, seyit.camtepe\}@data61.csiro.au}}
		\authorrunning{Thapa et al.}
		

		%
		%
		\titlerunning{Authors' Preprint}
		
		\maketitle
		

		\abstract{
			%
			In the distributed collaborative machine learning (DCML) paradigm, federated learning (FL) recently attracted much attention due to its applications in health, finance, and the latest innovations such as industry 4.0 and smart vehicles.  
			FL provides privacy-by-design. It trains a machine learning model collaboratively over several distributed clients (ranging from two to millions) such as mobile phones, without sharing their raw data with any other participant.
			In practical scenarios, all clients do not have sufficient computing resources (e.g., Internet of Things), the machine learning model has millions of parameters, and its privacy between the server and the clients while training/testing is a prime concern (e.g., rival parties). In this regard, FL is not sufficient, so split learning (SL) is introduced. SL is reliable in these scenarios as it splits a model into multiple portions, distributes them among clients and server, and trains/tests their respective model portions to accomplish the full model training/testing. In SL, the participants do not share both data and their model portions to any other parties, and usually, a smaller network portion is assigned to the clients where data resides. 
			Recently, a hybrid of FL and SL, called splitfed learning, is introduced to elevate the benefits of both FL (faster training/testing time) and SL (model split and training). 
			Following the developments from FL to SL, and considering the importance of SL, this chapter is designed to provide extensive coverage in SL and its variants. The coverage includes fundamentals, existing findings, integration with privacy measures such as differential privacy, open problems, and code implementation.   
			%
		}
		

		\section{Introduction}
		\label{sec:introduction}
		
		In today's world, machine learning (ML) has become an integral part in various domains, including health~\cite{health,reviewpaper_our}, finance~\cite{finance} and transportation~\cite{transportation}. As data are usually distributed and stored among different locations (e.g., data centers and hospitals), distributed collaborative machine learning (DCML) is used over conventional (centralized) machine learning to access them without centrally pooling all data. Moreover, additional advantages of DCML over conventional machine learning includes improved privacy of data by limiting it within its primary locations and distributed computations. Overall, DCML has become an essential tool for training/testing ML/AI models in a wide scale of distributed environments ranging from few to millions of devices~\cite{fedsurvey}.    
		
		DCML introduces collaborative architectures that enable reliable solutions for data sharing and model training~\cite{kraska2013mlbase}. Federated learning (FL) is a DCML mechanism that provides a widely accepted approach for ML/AI. In FL, several distributed clients (e.g., mobile phones)  train a global ML/AI model collaboratively in an efficient and privacy-preserving manner by keeping data within the local boundaries (i.e., data custodians/clients) while sharing no data among participating entities~\cite{yang2019federated}. Moreover, in FL, each client keeps a copy of the entire model and trains it with the local data, and in a subsequent step, the local models (trained locally) are forwarded to a coordinating server for the aggregation, which produces a global model. However, in a resource-constrained environment (e.g., the internet of things environments) where the clients have limited computing capability, FL is not feasible. As the server is assumed to have high computing resources than clients, FL is not leveraging that as the sever performs low computing jobs such as model aggregation and coordination. In contrast, the server can compute a larger portion of the ML model and reduce the computation at the client-side. As such, split learning (SL) is introduced.    
		
		SL is a DCML approach that introduces a split in executing a model training/testing that is shared among clients and the server~\cite{split_differentconfiguration}. The server and clients have access only to their portion of the whole model. Thus, it provides model privacy while training/testing in contrast to FL; in addition to keeping input data in the local bounds~\cite{split_differentconfiguration}. Besides, SL can be communication efficient and achieve faster convergence than FL~\cite{comm_efficiency}. Due to these inherent features of SL, it is gaining much research attention. There have been various research works performed to address its multiple aspects, including performance enhancement by introducing a variant of SL, called splitfed learning (SFL)~\cite{splitfed}, leakage reduction~\cite{split_leakage1}, and communication efficiency~\cite{comm_efficiency}. However, there are many open problems, including efficient leakage reductions, handling non-IID data distributions among clients, and reducing communication costs. An extensive understanding of SL is required to address these problems.
		
		This chapter provides comprehensive knowledge on SL and its variants, recent advancements, and integration with privacy measures such as differential privacy and leakage reduction techniques. In this regard, this chapter is divided into four main sections: (S1) federated learning to split learning, (S2) data privacy and privacy-enhancing techniques, (S3) applications and implementation, and (S4) challenge and open problems. 
		Firstly, Section S1 presents the fundamentals, dynamics, crucial results produced under FL, and detailed coverage on SL and SFL. The primary focus is given to SL. Detailed coverage is done on the key results that include performance analysis, effects of the number of clients on the performance, communication efficiency, data privacy leakage, and methods to perform SL over partitioned datasets (e.g., vertical partitioned) among clients.   
		Secondly, Section S2 explores the dynamics of privacy offered by FL, SL, and SFL in their default settings. Afterward, this chapter discusses privacy measures such as encryption-based techniques and provides details on the provable-privacy mechanism, called differential privacy, which are employed further to enhance the data privacy with these DCML techniques. 
		Thirdly, Section S3 presents the significance of SL and SFL by highlighting their applications in various domains, including health. Besides, it provides technical background on implementing the related concepts highlighting programmatic techniques with code examples. 
		Finally, open challenges and potential future research directions are discussed in Section S4.

		\section{Federated learning to split learning}
		In this section, firstly, FL is revisited to highlight its relevant results for completeness. However, this chapter does not intend to conduct an in-depth analysis of FL as its primary goal is to investigate the architectural improvements of FL towards SL and SFL. Hence, the discussions on FL act as a base to the rest of the chapter. Consequently, the discussions on FL is followed by an in-depth analysis of SL and its variants by covering their existing results and recent developments.
		\subsection{Overview of federated learning and key results}
		\label{sec:fed_overviewNresults}
\begin{figure}[tbh]
			\centering
			\includegraphics[trim=8cm 5cm 8cm 3.5cm, clip=true, width=0.35\linewidth]{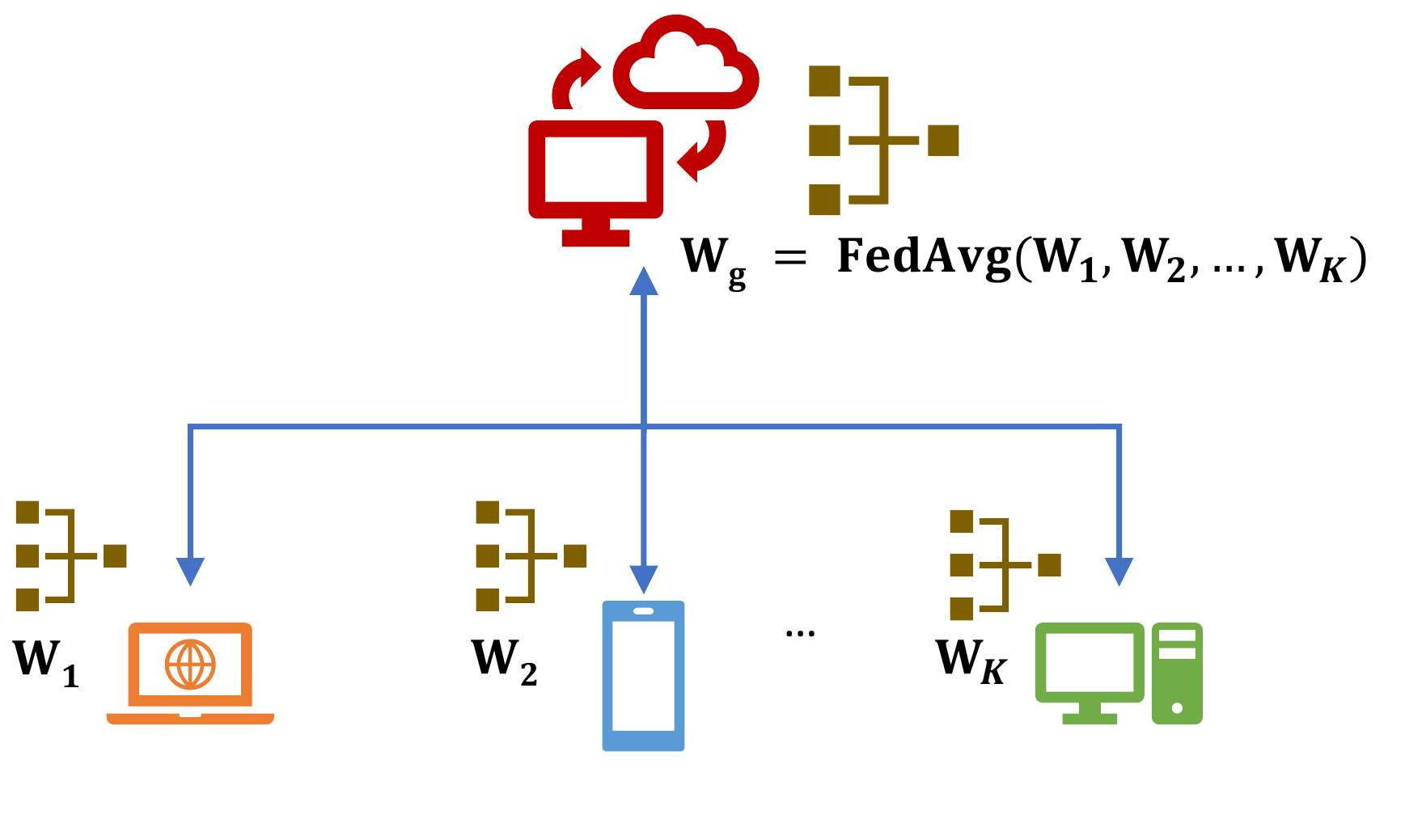}
			\caption{Federated Learning with $K$ clients.}
			\label{fig:fed_learning}      
		\end{figure}

		The primary concept behind FL is training an ML/AI model without exposing the raw data to any participant, including the coordinating server, enabling the collaborative training/testing of the model between distributed data owners/custodians~\cite{yang2019federated}. Fig.~\ref{fig:fed_learning} illustrates FL with $K$ clients. In FL, the model knowledge orchestration is done usually through federated averaging ({FedAvg})~\cite{fed1}.  
		
		Due to the promising aspects of the underlying architecture, FL is gaining much attention in privacy-preserving data analytics. Google and Apple are two major companies that utilize FL capabilities in training ML/AI models~\cite{hard2018federated,fedsurvey}. Besides, it is gaining attention in transportation (e.g., self-driving cars) and Industry 4.0~\cite{arachchige2020trustworthy,liu2020privacy}. Several implementation-based frameworks are already supporting FL. Some of these frameworks include  PySyft~\cite{ryffel2018generic}, Leaf~\cite{caldas2018leaf}, and PaddleFL~\cite{he2020fedml}.

		
		%
		
		Based on different requirements, such as the feature space distribution, different FL configurations can be employed. These configurations include (1) horizontal federated learning, (2) vertical federated learning, and (3) federated transfer learning~\cite{yang2019federated}. Horizontal federated learning is used when all distributed clients share the same feature space but different data samples. Vertical federated learning is used when the distributed clients have different feature spaces that belong to the same data samples, and federated transfer learning is used when the distributed clients have different datasets that have different feature spaces belong to different data samples~\cite{yang2019federated}. However, one of the fundamental issues in FL is the communication bottleneck. When clients need to connect through the Internet, communications become potentially expensive and oftentimes unreliable. Besides, the limited computational capabilities of distributed clients often make the whole process slower in model training. FL requires a particular client to hold the entire model. Hence, in a situation where a heavy deep learning model needs to be trained, an FL client should have enough computing power, which avoids resource-constrained devices being employed~\cite{li2020federated}. Besides, the requirement of continuous communications with the server also limits FL's use in a conventional environment that does not possess powerful communication channels. Moreover, in real-world scenarios, distributed nodes/clients can face different failures, directly affecting the global model generalization. Hence, the participation of an extensively large number of clients can make FL unreliable. Different compression mechanisms such as gradient compression, model broadcast compression, and local computation reduction are being experimented to maintain an acceptable FL efficiency~\cite{zhao2018federated,haddadpour2020federated}. It has also been shown that unbalanced and non-IID data partitioning across unreliable devices drastically affects model convergence~\cite{kairouz2019advances}. Data leakage introduces another layer of complexity for FL in addition to such data communication and data convergence issues. It has been shown that the parameters transferred between the clients and the server can leak information~\cite{aono2017privacy}. The participation of malicious clients or servers can introduce many security and privacy vulnerabilities—the backdoor attacks is an example where malicious entities can deploy backdoor to learn others' data~\cite{bagdasaryan2020backdoor}. Besides, privacy attacks, such as membership inference, can exploit this vulnerability~\cite{shokri2017membership}. Many existing works try to employ third-party solutions such as fully homomorphic encryption and differential privacy to limit such unanticipated privacy leaks from FL. However, the performance of these approaches is always governed by the limitations of these approaches (e.g., high computational complexity~\cite{acar2018survey}). In a distributed setup where resource-constrained devices are used, this can be a challenging task. 

		\subsection{Split learning and key results}
		In this section, firstly, SL is introduced along with its algorithm and configurations. Secondly, an extensive coverage on its existing results, including its performance in model development, data leakage analysis, and countermeasures, are presented. 
		
		\subsubsection{Split learning}
		
		\begin{figure}[H]
			\centering
			\includegraphics[trim=1.5cm 0cm 1.5cm 1cm, clip=true, width=0.35\linewidth]{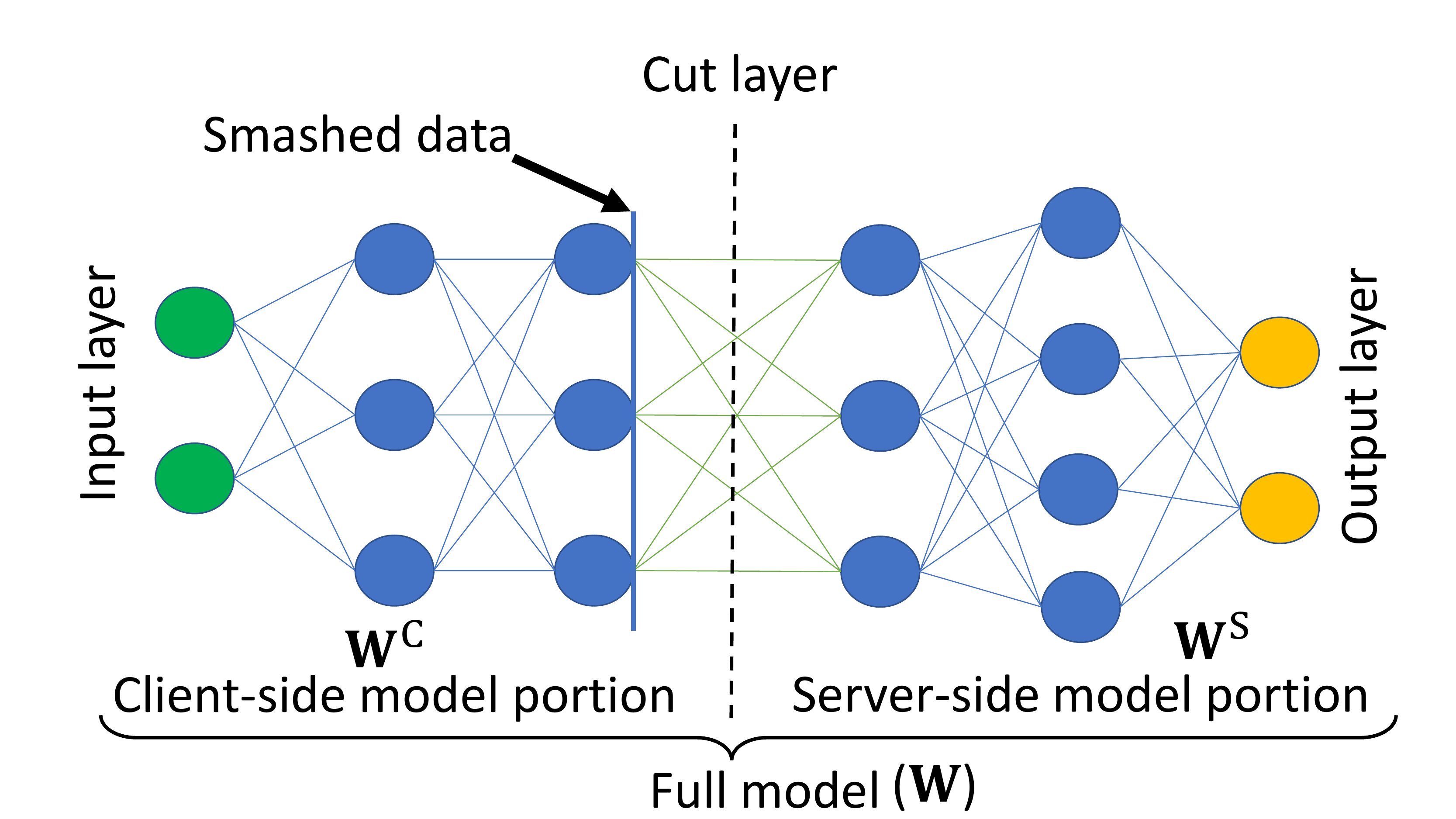}
			\caption{Split learning in a simple setup.}
			\label{fig:sec_split_fig1}      
		\end{figure}
		
		SL~\cite{Split_firstpaper,split_differentconfiguration} is a collaborative distributed machine learning approach, where an ML network (model) is split into multiple portions and executed in sequence at different clients and servers. In a simple setup, the full model $\mathbf{W}$, which includes weights, bias and hyperparameters, is split into two portions $\mathbf{W}^{\textup{C}} $ and $\mathbf{W}^{\textup{S}}$; $\mathbf{W}^{\textup{C}} $ is called client-side network portion and $\mathbf{W}^{\textup{S}}$ is called server-side network portion, as illustrated in Fig.\ref{fig:sec_split_fig1}. In SL, the client commits only to the training/testing of the client-side network, and the server executes only to the training/testing of the server-side network. The training and testing of the full model are done by executing a sequential (forward/backward) propagation between a client and the server. In the simplest form, firstly, the forward propagation takes place as follows: a client forward propagates until a specific layer of the network, which is called the \emph{cut layer}, over the raw data, then the cut layer's activations, called \emph{smashed data}, are transmitted to the server. Smashed data matrix is represented as $\mathbf{A}$. Afterward, the server considers the smashed data (received from the client) as its input and performs the forward propagation on the remaining layers. So far, a single forward propagation on the full model is completed. 
		Now the back-propagation takes place as follows: After calculating the loss, the server starts back-propagation, where it computes gradients of weights and activations of the layers until the cut layer, and then transmits the smashed data's gradients back to the client. With the received gradients, the client executes its back-propagation on its client-side network. So far, a single pass of the back-propagation between a client and the server is completed. 
		In ML/AI model training, the (forward and back) propagation continue until the model is trained on all the participating clients and meet a decent convergence point (e.g., high prediction accuracy). For details on SL, refer to Algorithm~\ref{alg:split} (which is extracted from~\cite{splitfed}).  
		
		By limiting the client-side model network portion up to a few number of layers, SL reduces the client-side computation compared to FL, where each client has to train a full-sized model. 
		Besides, while machine learning training/testing, the server and clients are limited within their designated portions of the full model, and the full model is not accessible to them. To predict the missing model portion, a semi-honest\footnote{A semi-honest entity in collaboration among multiple entities performs their jobs as specified, but it can be curious about the details of the information present in other participating entities. It is also called an honest-but-curious entity.} client or server requires to predict entire parameters of the missing model portion, and its probability decrease with the increase in model parameters in the portion. As a result, SL provides a certain level of privacy to the trained model from honest but curious clients and the server. There is no model privacy between the clients and server in FL as both of them have either white-box access (i.e., full access) to the full model, or the server can easily predict the full model from the gradients of the locally trained model that are transferred by clients for model aggregation. 
		
		\begin{figure}[!htb]
			\centering
			\includegraphics[trim=5cm 0.5cm 5cm 1cm, clip=true, width=0.35\linewidth]{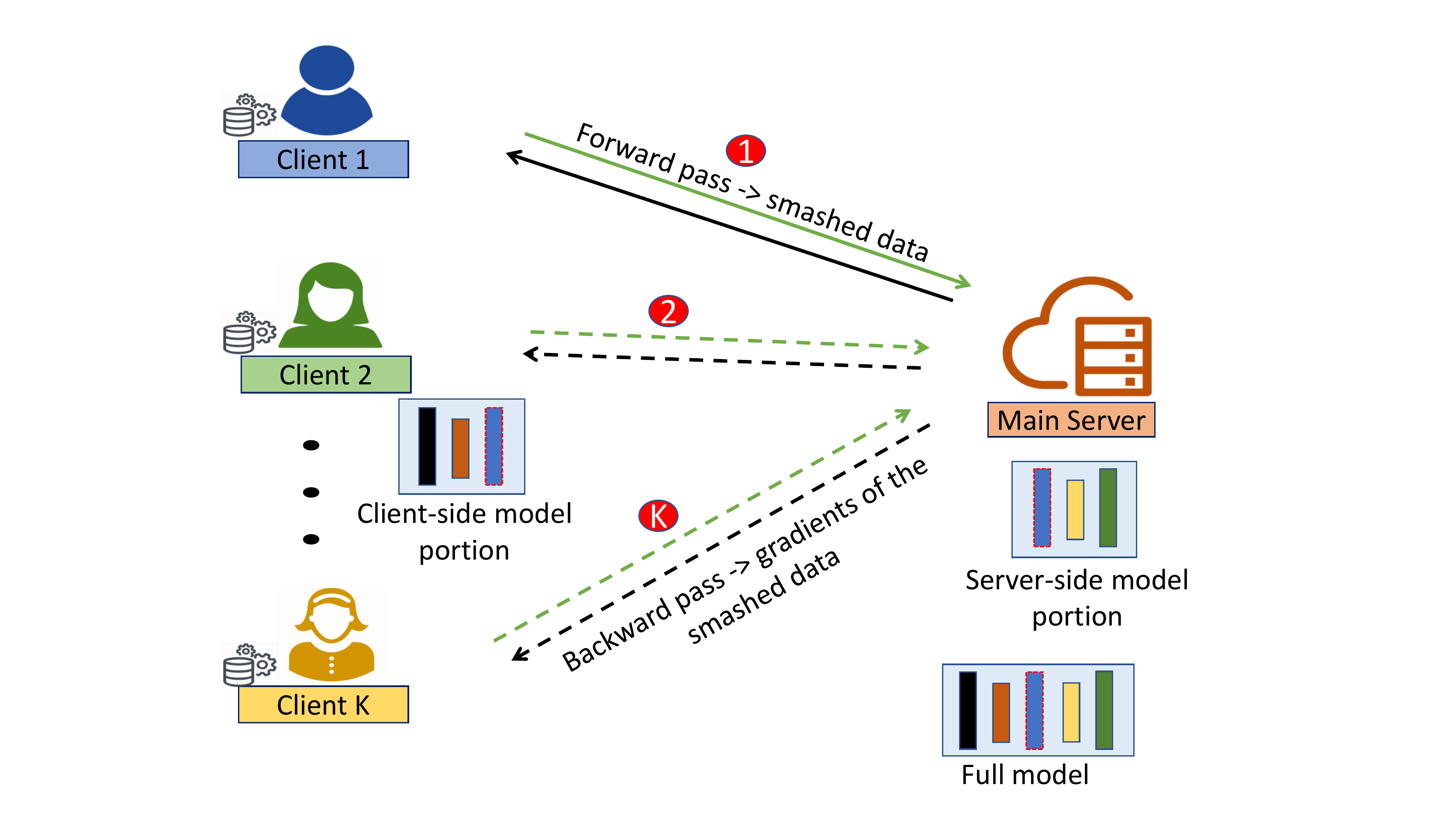}
			\caption{Split learning with multiple clients. At one round, indicated by the red dot with a label, one client $k\in \{1,2,\dotsc,K\}$ engages with the server for the training/testing. The control is passed to client $k+1$, till $k+1 = K$, after the client $k$ completes its operations with the server.}
			\label{fig:sec_split_fig2}      
		\end{figure}
		
	
		\begin{algorithm} [!htb]
			
			\small
			\SetNoFillComment
			\caption{\small Split learning with label sharing~\cite{split_differentconfiguration,splitfed}.} 
			\SetAlgoNoLine
			%
			\textbf{Notations:}
			\begin{itemize}
				\item $S_t$ is a set of $n_t$ clients at time instance $t$.
				\item $\mathbf{A}_{k,t}$ is the smashed data of client $k$ at $t$.
				\item $\mathbf{Y}_{k}$ and $\hat{\mathbf{Y}}_{k}$ are ground truth and predicted labels for client $k$.
				\item $\triangledown\ell$ is the gradient of loss $\ell$.
				\item $\eta$ is the learning rate.
			\end{itemize}
			\vspace{4pt}
			\SetKwProg{Fn}{EnsureMainServer executes at time instance $t\geq 0$:}{}{} \tcc{	\scriptsize Runs on Server}
			\Fn{}{
				
				\For{\textup{a request from client $ k\in S_t $ with new data}} {
					$ (\mathbf{A}_{k,t}, \mathbf{Y}_{k}) \leftarrow$ ClientUpdate$(\mathbf{W}^{\textup{C}}_{k,t})$ \quad
					\Comment{\emph{$ \mathbf{A}_{k,t} $ and $ \mathbf{Y}_{k} $ are from client $k$}}	\\
					Forward propagation with $ \mathbf{A}_{k,t} $ on $ \mathbf{W}^{\textup{S}}_t$\quad  
					\Comment{\emph{$\mathbf{W}^{\textup{S}}_t$ is the server-side part of the model $ \mathbf{W}_t $}}\\
					Loss calculation with $ \mathbf{Y}_{k}$ and  $\hat{\mathbf{Y}}_{k}$ \\
					Back-propagation and model updates with learning rate $ \eta $:
					$ \mathbf{W}^{\textup{S}}_{t+1}  \leftarrow \mathbf{W}^{\textup{S}}_t - \eta\ \triangledown \ell (\mathbf{W}^{\textup{S}}_t; \mathbf{A}^{\textup{S}}_t)$\\
					Send $ d\mathbf{A}_{k,t} := \triangledown\ell (\mathbf{A}^{\textup{S}}_t; \mathbf{W}^{\textup{S}}_t) $ (i.e., gradient of the $ \mathbf{A}_{k,t} $) to client $ k $ for its ClientBackprop$  (d\mathbf{A}_{k,t}) $\\
			}}
			\vspace{5pt}
			\SetKwProg{Fn}{EnsureClientUpdate$ ( \mathbf{W}^{\textup{C}}_{k,t}) $:}{}{} \tcc{	\scriptsize Runs on Client  $k \in \{1,\dotsc,K\}$ }	
			\Fn{}{
				Set $ \mathbf{A}_{k,t} $ = $ \phi $ \\
				\eIf{\textup{Client $k$ is the \emph{first} client to start the training}}{$\mathbf{W}^{\textup{C}}_{k,t}\leftarrow$ Randomly initialize (using Xavier or Gaussian initializer)  }{
					$\mathbf{W}^{\textup{C}}_{k,t} \leftarrow$ ClientBackprop$(d\mathbf{A}_{k-1,t-1})$\quad \Comment{\emph{$k-1$ is the last trained client with the main server}}
				}
				
				\For{\textup{each local epoch $ e$ from $ 1 $ to $E$}}{
					\For{\textup{batch} $ b \in  \mathcal{B}$ }{
						Forward propagation on $ \mathbf{W}^{\textup{C}}_{k,b,t} $ 
						\Comment{\emph{$\mathbf{W}^{\textup{C}}_{k,t}$ of batch $ b $, where $\mathbf{W}^{\textup{C}}_{k,t}$ is the client-side part of the model $ \mathbf{W}_t $}}\\
						Concatenate the activations of its final layer to $ \mathbf{A}_{k,t} $\\
						Concatenate respective true labels to $  \mathbf{Y}_{k}$\\ 
					}
				}
				Send $ \mathbf{A}_{k,t} $ and $\mathbf{Y}_{k}$ to server\\}
			\vspace{5pt}	
			\SetKwProg{Fn}{EnsureClientBackprop$  (d\mathbf{A}_{k,t}) $:}{}{} \tcc{	\scriptsize Runs on Client $ k$}	
			\Fn{}{
				\For{\textup{batch} $ b \in  \mathcal{B}$}{
					Back-propagation with $ d\mathbf{A}_{k,b,t} $\quad \Comment{\emph{$ d\mathbf{A}_{k,t} $ of the batch $ b $} }\\
					Model updates $ \mathbf{W}^{\textup{C}}_{t+1}  \leftarrow \mathbf{W}^{\textup{C}}_t - \eta\ d\mathbf{A}_{k,b,t} $\\	
				}
				Send $ \mathbf{W}^{\textup{C}}_{t+1}$ to the next client ready to train with the main server.
			}
			\label{alg:split}
		
		\end{algorithm}

		
		\paragraph{\textbf{Split learning with multiple clients:}}
		\label{sec:split_with_multipleclients}
		SL with multiple clients is illustrated in Fig.~\ref{fig:sec_split_fig2}. With multiple clients, SL takes place in two ways, namely centralized distributed training and peer-to-peer distributed training~\cite{Split_firstpaper}. In centralized distributed training, after a client completes its training with the main server, it uploads an encrypted version of the weights of the client-side network either to the server or a third-party server. When a new client initiates its training with the server, it first downloads the encrypted weights, decrypts them, and loads them to its client-side network. In peer-to-peer distributed training, the server sends the address of the last trained client to the new client, which then downloads the encrypted weights directly from the client and loads them to its client-side network after decryption.
		Besides, training can be carried out in two ways; one with the client-side model synchronization (client $k$ trains then the model is passed to client $k+1$), and the other is without weight synchronization, where clients take turns with alternating epochs in working with the server. However, there is no convergence guarantee if model training is done without weight synchronization.
		
\begin{figure}[t]
			\centering
			\subfigure[]{
				\includegraphics[trim=10cm 0.5cm 9cm 0.5cm, clip=true, height=0.2\columnwidth]{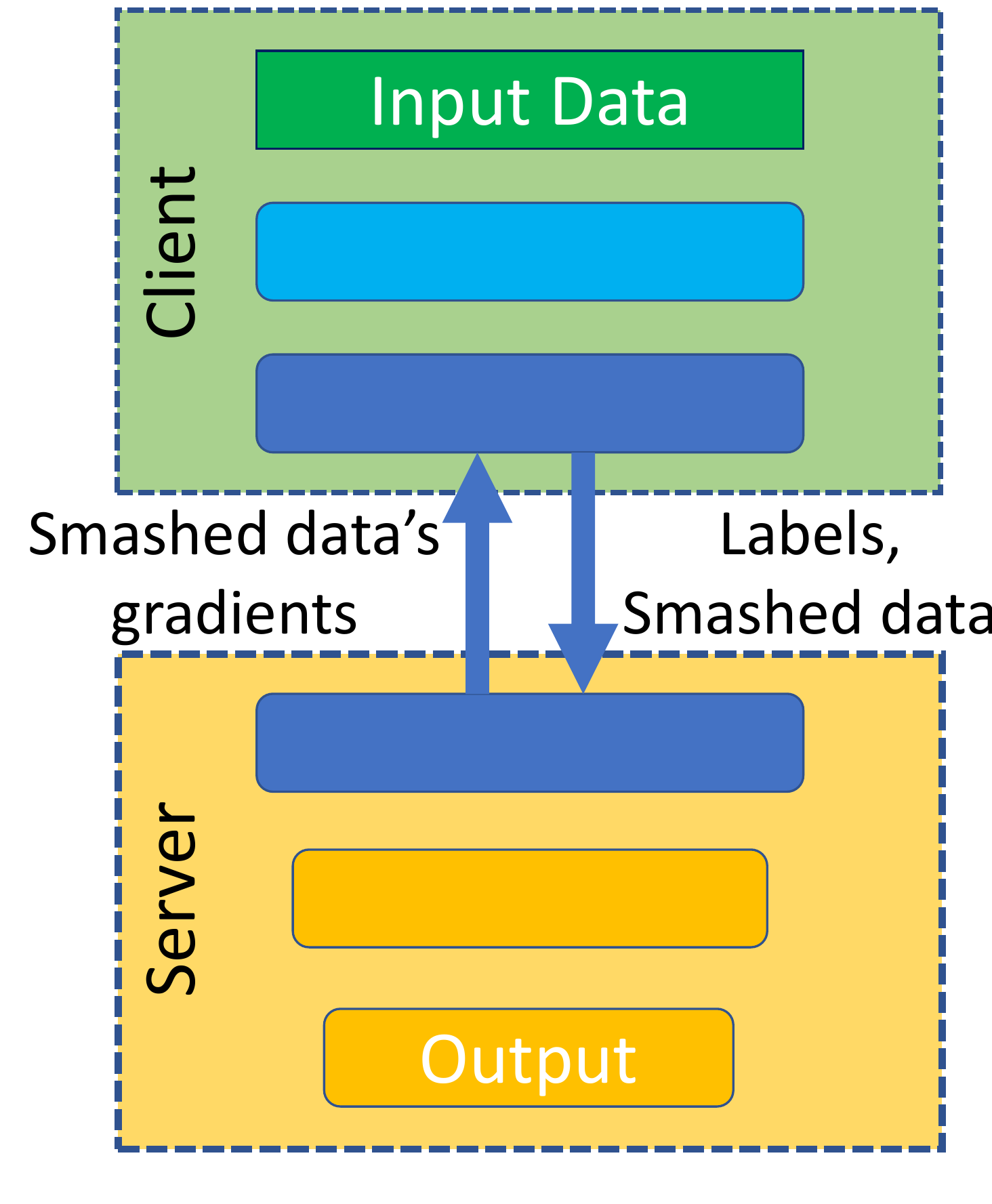}
			}
			\hskip-5pt
			\subfigure[]{
				\includegraphics[trim=8cm 0.5cm 7cm 0.5cm, clip=true, height=0.2\columnwidth]{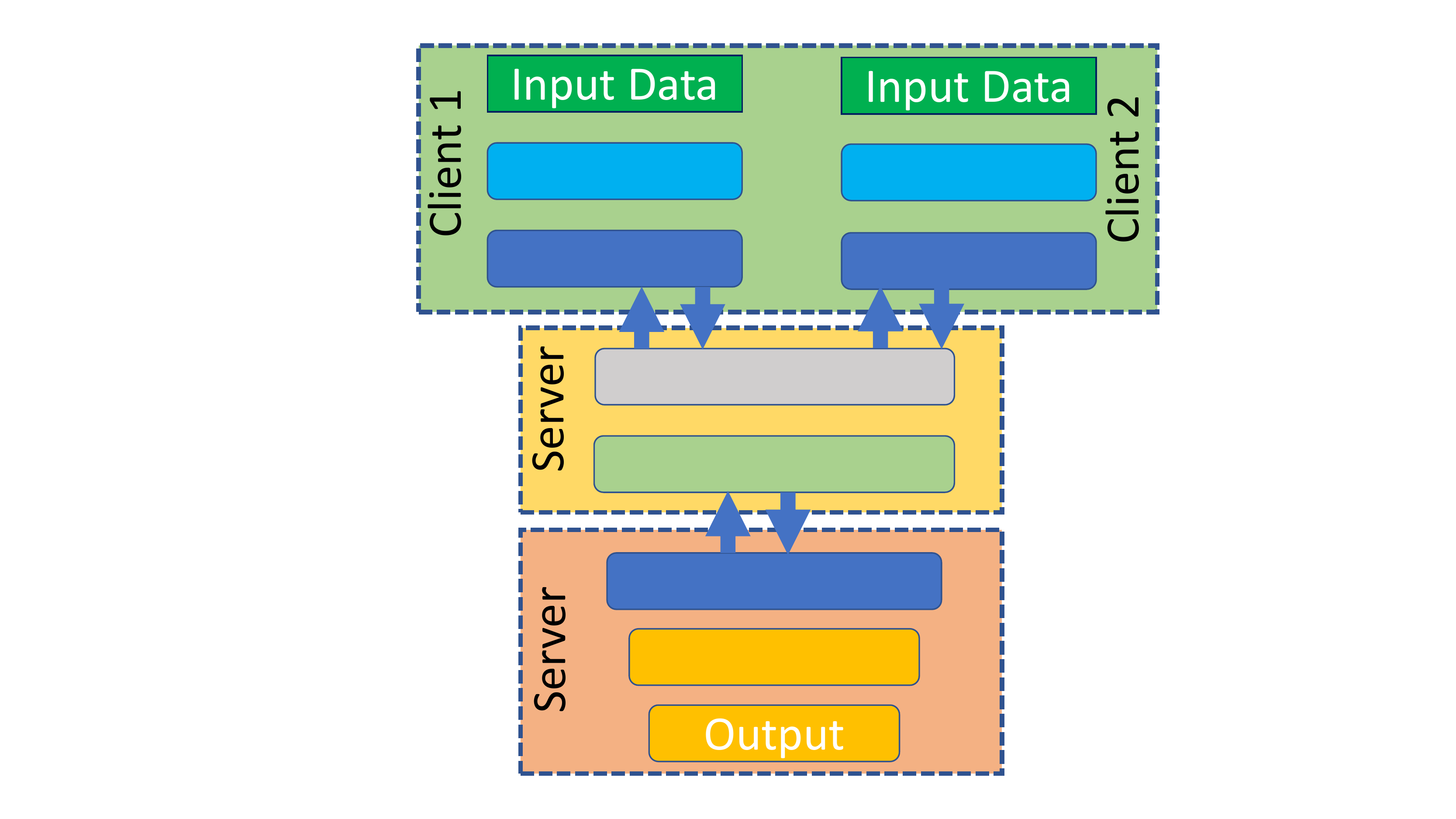}
			}
			\hskip-5pt
			\subfigure[]{
				\includegraphics[trim=7cm 1cm 7cm 1cm, clip=true, height=0.2\columnwidth]{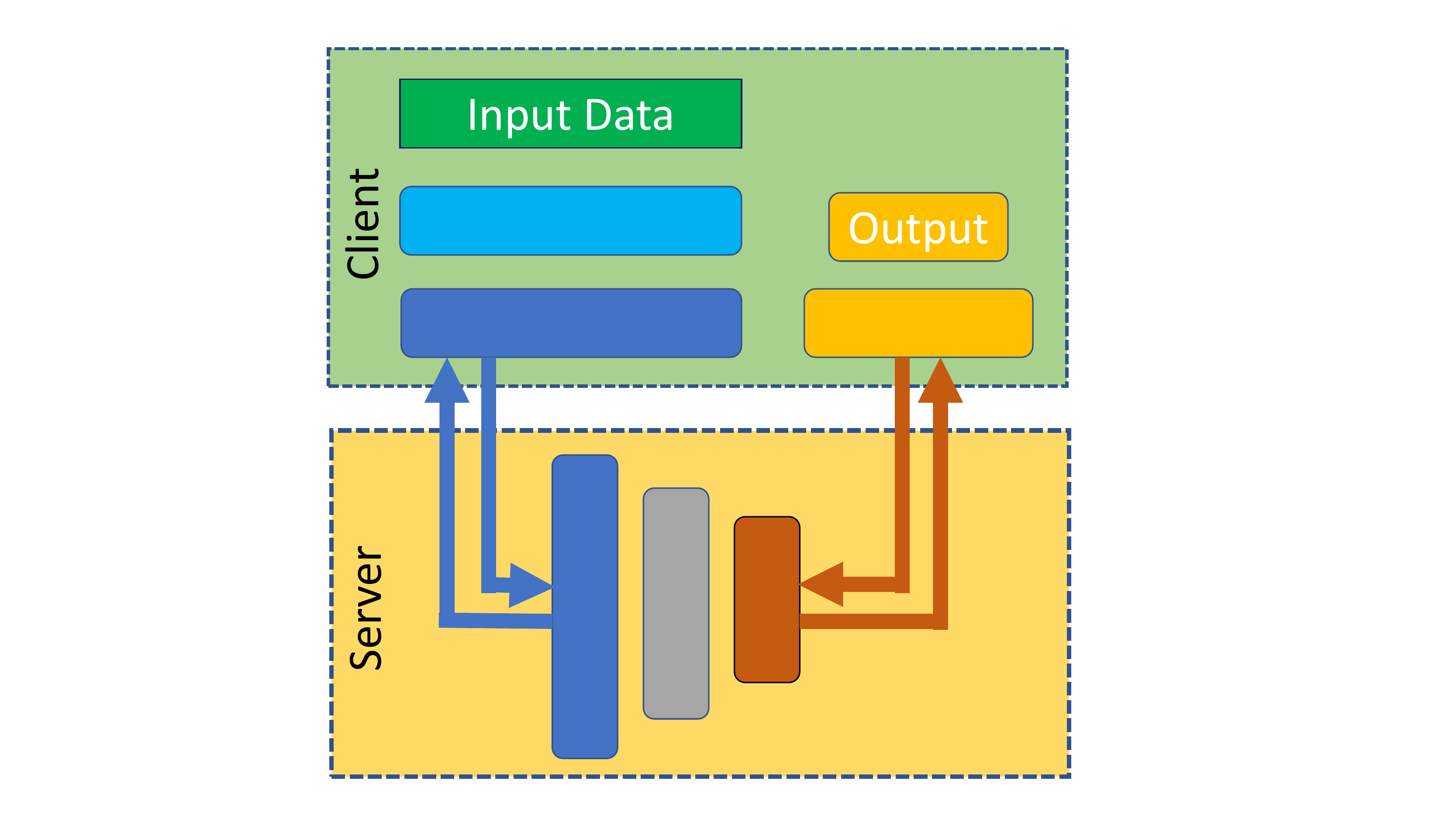}
			}
			\hskip-5pt
			\subfigure[]{
				\includegraphics[trim=8cm 0.5cm 5.5cm 0.5cm, clip=true, height=0.2\columnwidth]{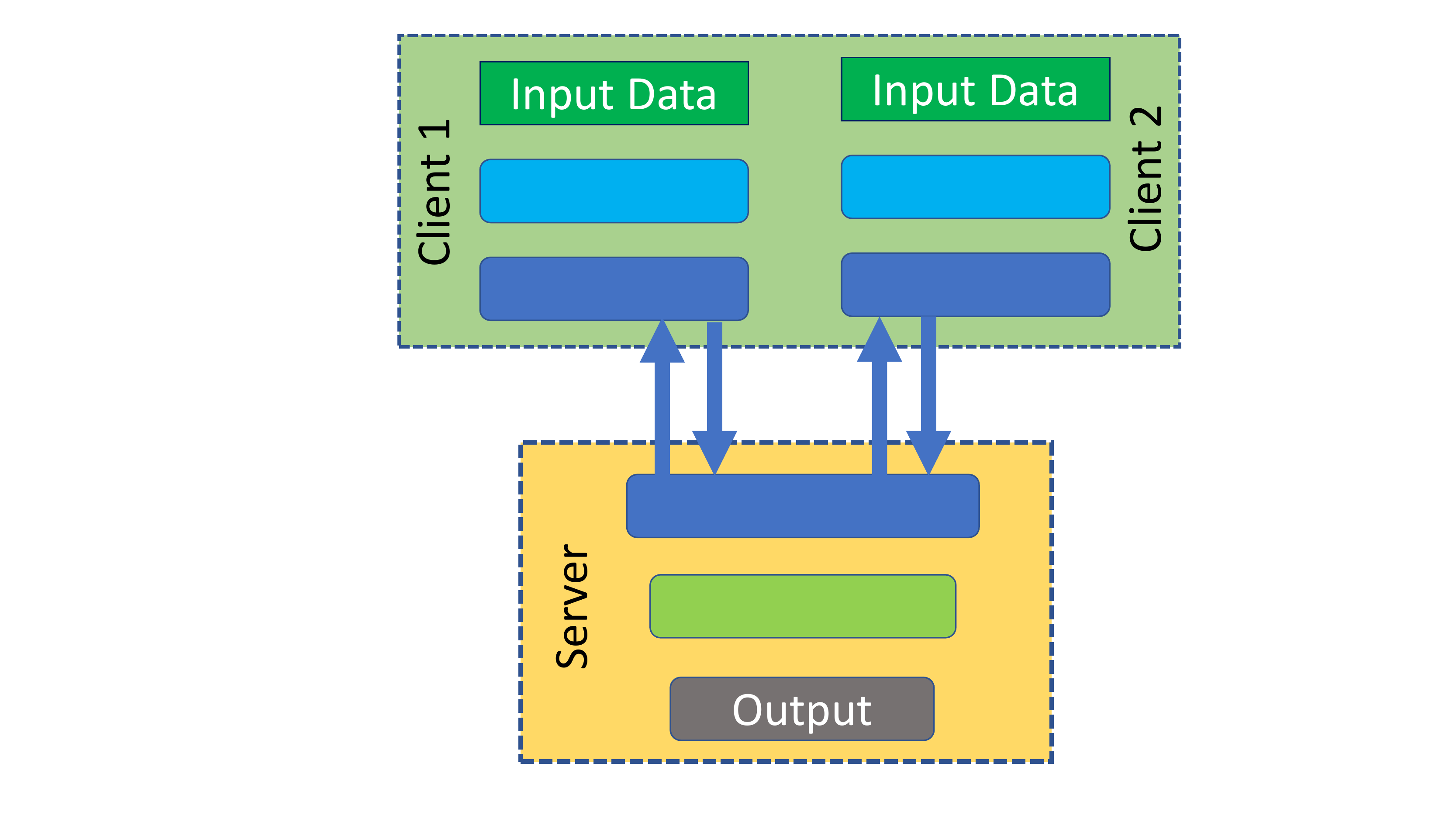}
			}
			\caption{Split learning configurations (a) simple vanilla, (b) extended vanilla, (c) without label sharing, and (d) vertically partitioned data.}
			\label{fig:diffconfiguration}
		\end{figure}

		\paragraph{\textbf{Different configurations in split learning:}}
		\label{sec:diffconfsplit}
		Due to the flexibility of splitting the model while training/testing, SL has several possible configurations, namely \emph{vanilla split learning}, \emph{extended vanilla split learning}, \emph{split learning without label sharing}, \emph{split learning for a vertically partitioned data}, \emph{split learning for multi-task output with vertically partitioned input}, \emph{`Tor' like multi-hop split learning}~\cite{split_differentconfiguration}. Refer to Fig.~\ref{fig:diffconfiguration} for illustrations.

		Vanilla split learning is the most straightforward configuration, in which the client shares the smashed data and labels with the server (Fig.~\ref{fig:diffconfiguration}(a)). In an extended vanilla split learning, some other workers process some intermediate layers of the ML network before passing it to the main server (Fig.~\ref{fig:diffconfiguration}(b)). An analogous extension of the vanilla configuration provides the `Tor' like multi-hop split learning configuration, where one client has data, and multiple clients/servers train a portion of the ML network in a sequence. 
		In SL without label sharing configuration, the client only shares the smashed data to the server (unlike vanilla split learning), which completes forward propagation up to some layers of the network, called \emph{server cut layer}, and then again sends the activations of its server cut layer to the client, which then completes the forward propagation up to the output layer. Afterward, the client starts backpropagation and sends the gradients of the activations of the server cut layer to the server. Then the server carries out the backpropagation and sends the smashed data's gradients to the client, which then completes its backpropagations. Here the forward propagation and backpropagation happen in a U-shape (client $\Longleftrightarrow$ server $\Longleftrightarrow$ client), thus it is also called U-shaped configuration (Fig.~\ref{fig:diffconfiguration}(c)). 
		For SL with vertically partitioned data configuration (Fig.~\ref{fig:diffconfiguration}(d)),  clients have a vertically partitioned dataset configuration, and each carries out the forward propagation on its local client-side ML network portions. Then they transfer their smashed data to the server. Afterward, the server concatenates all smashed data and carry on the forward propagation on the single server-side model. The backpropagation proceeds from the output layer up to the concatenation layer at the server, which then transmits the respective smashed data's gradients to the clients. Next, the clients perform their backpropagation on their client-side ML network portions.
		In SL for multi-task output, multi-modal data from different clients train their client-side ML network up to their corresponding cut layer and then transfer their smashed data to an intermediate agent, which concatenates the smashed data from all clients. Afterward, it sends the concatenated smashed data to multiple servers, where each server trains its server-side model. 
		
		All these configurations in SL are useful based on the requirements. For example, if labels are sensitive to the clients, then there is no need to send it to the server; instead, U-shape SL can be used as the learning configuration. If there is a need to keep the identity of clients confidential to the server, then it can be done by the extended vanilla configuration or tor configuration.
		
		\subsubsection{Key results in split learning}
		There are several lines of works in SL literature. Broadly, these works are divided into five main categories, namely convergence, computation, communication, dataset partition, and information leakage.
		
		\paragraph{\textbf{Convergence:}} 
		\label{sec:convergence}
		This category explores and addresses the research questions related to the convergence in SL; the effect on the model convergence due to the data distribution among participating clients and their number. The data distribution is of two types: independent and identically distributed (IID) and non-IID. In IID distribution, the training datasets distributed among the clients are IID  sampled from the total dataset, where each sample has the same probability distribution and mutually independent of each other. The non-IID dataset distribution in a distributed setup with multiple clients refers to the difference in the distribution and any dependencies of the local datasets among the clients. The non-IID dataset due to the quantity skew (i.e., different number of samples in different clients), and label distribution skew (i.e., different clients can have samples belonging to the particular labels) are explored in SL. With a single client and server setup, SL has the same results as centralized learning, where the ML network is not split and trained~\cite{Split_firstpaper}. 
		
		Under IID data configurations, SL shows higher validation/test accuracy and faster convergence when considering a large number of clients than FL~\cite{Split_firstpaper,ourpaper1}. The experiments are carried out with models, including VGG\footnote{VGG16 has sixteen network layers, and its input dimension of an image dataset is  $224\times 224\times 3 $ and $3 \times 3$ sized kernels are used.}~\cite{vgg16} and ResNet\footnote{ResNet18 has eighteen network layers, and a $ 3\times 3$ and $ 7\times 7 $ sized kernels are implemented in its layers.}~\cite{resnet}, over several datasets, including MNIST dataset (a database with 70,000 handwritten digit samples of ten labels), CIFAR-10 dataset (60,000 tiny images with ten labels), and ILSVRC-12 (1.2 million image samples with 1000 object categories), speech command (SC) dataset (20,827 samples of single spoken English word with ten classes), and electrocardiogram (ECG) dataset (26,490 samples representing five heartbeat types via ECG signals). The number of clients in SL affects the convergence curve. It fluctuates more as the number of clients goes up~\cite{ourpaper1}. For AlexNet\footnote{AlexNet has eight network layers, and its layers has $ 3\times 3$, $5\times 5$, and $ 11\times 11 $ sized kernels. The dimension of the input image is $ 227\times 227\times 3 $.}~\cite{alexnet} on HAM10000~\cite{hamdata} (medical image dataset with 10,015 samples with seven cases of skin lesion), it even fails to converge for 100 clients~\cite{splitfed}.
		\makeatletter{\renewcommand*{\@makefnmark}{}
			\footnotetext{VGG16, ResNet18, and AlexNet are convolutional neural networks with multiple layers. The layers has convolution, ReLU (activation function), max pooling (down sampling), and fully connected layers.}\makeatother}
		
		For imbalanced data distributions (different sample sizes in different clients ranging from 48 to 3855), the convergence of FL is slow if there is an increase in the number of clients (50 and 100). In contrast, for the same setup, SL's fast convergence shows less sensitivity in convergence towards imbalanced data distributions. However, the model performance (e.g., accuracy) decreases if there is an increase in the number of clients for the case with imbalance data distribution~\cite{ourpaper1}. 
		SL is highly sensitive to non-IID data distributions than FL. For ECG and SC dataset, SL is slow in convergence and no convergence if each client has samples from only one class~\cite{ourpaper1}. In experiments performed in~\cite{ourpaper1}, FL outperforms in accuracy and convergence than SL under a non-IID setting. Overall, it is required to carefully choose an appropriate DCML based on the dataset distribution and the number of clients.

		\paragraph{\textbf{Computation:}}
		This category explores and addresses the research questions related to the client-side computational efficiency due to SL and the techniques for improvements by maintaining data privacy. Due to the inherent characteristic of SL, i.e., split the network and train them separately on clients and server, it reduces the computation significantly in the client-side~\cite{Split_firstpaper}. It is shown that for a setup with 100 and 500 clients, when training CIFAR10 over VGG, SL requires 0.1548 TFlops\footnote{Floating point operations per second (Flops) is a measure of the computation of the instructions per second, and one Tera Flops (TFlops) is $10^{12}$ Flops.} and 0.03 TFlops of computations, respectively, in the client-side. On the other hand, FL requires 29.4 TFlops and 5.89 TFlops, for the same setup, respectively~\cite{split_differentconfiguration}. The computation at the client-side depends on the client-size network size; however, how the split layer position affects the performance and the definition of the optimal position of the cut layer in a given model are still open problems. Moreover, the possible answers depend on other factors such as information leakage.

		\paragraph{\textbf{Communication:}}
		This category explores and addresses the research questions related to communication efficiency in SL and the techniques for further improvements. The communication cost in SL can be significantly less than FL by limiting the client-side network to few layers and few or compressed activations at the cut layer (e.g., max pooling layer)~\cite{Split_firstpaper}. In contrast, FL has gradient updates that of the full network from all clients to the main server, and global weights are forwarded from the server to all clients. For the same total dataset, the communication bandwidth of 6 GB and 1.2 GB per client are required for SL with 100 and 500 clients, respectively, with ResNet over CIFAR100. In contrast, FL requires 3 GB and 2.4 GB with 100 and 500 clients, respectively, for the same setup~\cite{split_differentconfiguration}.  
		
		\begin{table}[]
			\centering
			\caption{Communication efficiency~\cite{comm_efficiency}.}
			\begin{tabular}{|c|c|c|}
				\hline
				Method                                       & Communication per client & Total communication       \\ \hline
				Split learning with client weight sharing    & $2(p/K)q +\eta N$        & $2pq+\eta NK$ \\ \hline
				Split learning with no client weight sharing & $2(p/K)q$                & $2pq$                       \\ \hline
				Federated learning                           & $2N$                       & $2KN$                       \\ \hline
			\end{tabular}
			\label{tab:com_eff}
		\end{table}
		A more detailed analysis of the communication efficiency of SL and FL is done in~\cite{comm_efficiency}. The analytical results are presented in Table~\ref{tab:com_eff}. $K$ is the number of clients, $N$ is the number of model parameters, $p$ is the total dataset size, $q$ is the smashed layer's size, and $\eta$ is the fraction of the whole model parameters that belong to a client. Usually, the data are distributed, and the size of data at each client is $p/K$, where the size decreases if $K$ increases. 
		In a separate work~\cite{splitfed}, for ResNet18 on the HAM10000 dataset, SL is shown to have an efficient communication when the number of clients is twenty, whereas, for AlexNet on MNIST, it is efficient after five clients. These studies reconfirm the following result: In terms of communication efficiency, if the model size and the number of clients are large, then SL is efficient than FL; otherwise (for a small model and few clients), FL is efficient.

		{\bf{Note point:}} In SL, the communication cost also depends on the type of cut layer. The cut layer can be a max pool layer that compresses the outputs from the previous layer (e.g., convolutional layer) by $ \frac{n_{\textup{H}}} {\lfloor \frac{n_{\textup{H}}-f} {s} +1  \rfloor}$ times to the height and $ \frac{n_{\textup{W}}} {\lfloor \frac{n_{\textup{W}}-f} {s} +1  \rfloor}$ times to the width of the input dimension $n_{\textup{H}} \times n_{\textup{W}}$, where $f$ is the filter size of the pooling layer, and $s$ is the stride size. This is not the case if the cut layer is a convolutional layer, i.e., $n_\textup{H} \times n_\textup{W}$.
		
		\paragraph{\textbf{Dataset partition and split learning:}}
		This category explores and addresses the research questions related to carrying SL under various partitions of the datasets among the clients. Most of the works in SL consider the horizontal partition of the dataset, where different clients have different samples but all with the same feature space. However, in fields like finance, multiple clients (finance institutes) can have the same sample (related to the same set of data owners) but with different features. The dataset, in this case, is referred to as a vertically partitioned dataset.   
		
		SL with vertical partitioning of datasets is performed in~\cite{split_vertical}. This work evaluates several aggregating configurations to merge the outputs of the partial networks from multiple clients and analyze the performance and resource efficiency. The configurations include element-wise average, element-wise maximum, element-wise sum, element-wise multiplication, and concatenation. Considering three financial datasets, namely Bank Marketing, Some Credit Kaggle, and Financial PhraseBank, the experiments show that the performance depends on the dataset and the merging technique. Element-wise average pooling outperforms other techniques for Financial PhraseBank, whereas element-wise max pooling and concatenation for Bank Marketing and Give me Credit, respectively, outperforms other approaches. In other experiments, the performance degradation with the increase in the number of clients dropping during the training and testing is observed.
		Moreover, the communication and computation costs are measured for this setup, and their dependencies on the dataset and model architecture are observed. A similar technique of concatenating the smashed data from the clients before feeding to the server-side model is used in multiple classifications with horizontal partitioning of the dataset in SL~\cite{kim2020multiple}. This is done primarily for privacy reasons, as this setup does not require the client-side models to synchronize. However, a trusted worker to concatenate the smashed data is added to the setup.
		
		\paragraph{\textbf{Information leakage and countermeasures:}}
		\label{sec:inforleakage}
		This category explores and addresses the research questions related to privacy and information leakage and their countermeasures. Information leakage from an ML model is defined as the ability to reconstruct the original raw data from model parameters or intermediate activations. In SL, possible information leakage is an investigation based on the data communicated between the clients and the server. The server receives the smashed data from the clients in each epoch, and the smashed data can leak information about private/sensitive raw data as it possesses a certain level of correlation to the raw data.
		
		So far, in the literature, two methods are implemented to reduce information leakage in SL; one by using differential privacy (refer to Section~\ref{sec:diffprivacy} for details), and the other by mapping the smashed data with higher distance correlation~\cite{distance_correlation} to the raw input data. The latter is known to maintain accuracy, unlike differential privacy, which degrades accuracy with increased privacy. Moreover, it has relatively low computational complexity of $\mathcal{O}(n \textup{log} n)$ and $\mathcal{O}(n K \textup{log}n)$ for univariate and multivariate settings, respectively, with $\mathcal{O}(\textup{max}(n,k))$ memory requirements. $K$ is the number of random projections~\cite{split_leakage2}.
		Information leakage can be measured using Kullback-Leibler (KL) divergence between the raw data and the smashed data~\cite{split_leakage2,split_leakage1}. KL provides a measure of the invertibility of the smashed data. It is also known as relative entropy, and it is a non-symmetric measure of the difference between two probability distributions $X$ and $Z$. In other words, it is a measure of the information loss when $Z$ is used to approximate $X$ and expressed as follows:
		\begin{equation}
			D_{\textup{KL}}(X||Z) =  \sum^N_{i=1} X(x_i) \ln{\frac{X(x_i)}{Z(x_i)}}.
		\end{equation}
		KL divergence can be written as:
		\begin{equation}
			D_{\textup{KL}}(X||Z) =  H(X,Z) - H(X),
		\end{equation}
		where $H(X,Z)$ and $H(X)$ are the cross-entropy and entropy, respectively. The distance correlation is minimized by introducing an additional regularization term in the loss function:
		\begin{equation}
			\textup{Loss} = \alpha_1 \textup{DCOR} (X_n, \hat{Z}) + \alpha_2 \textup{CCE}(Y_n, \hat{Y}),
			\label{eq:loss}
		\end{equation}
		where DCOR refers to the distance correlation, CCE refers to categorical cross-entropy, and $\alpha_1$ and $\alpha_2$ are scalar weights. The combined optimization of the resulting loss term reduces the leakage from the smashed data without degrading accuracy. Furthermore, the same approach of distance correlation is used for attribute privacy, where the distance correlation is minimized between the smashed data and the certain attribute (e.g., age, race, and gender) of the raw data. For more details refer to~\cite{split_leakage2}.
		
		\subsection{Splitfed learning and key results}
		\label{splitfedsection}
		SL can be considered a better contender for edge-computing than FL due to application in a resource-constrained environment and model privacy. However, the client-side model synchronization between the clients is done sequentially, as described in Section~\ref{sec:split_with_multipleclients}. This results in a higher latency in the model training/testing in SL. Consequently, the latency adds challenges in SL to leverage the advantages of SL in a resource-constrained environment where fast model training/testing time is required to periodically update the model with the continually updating dataset with time. This environment characterizes the fields such as health and finance, where in a distributed setting, frequent model updates are required to incorporate newly available threats in activities, including real-time anomaly detection and fraud detection. In this regard, to utilize the benefits of both FL and SL, splitfed learning is introduced.
		
		\subsubsection{Splitfed learning}
		
		\begin{figure}[H]
			\centering
			\includegraphics[trim=1cm 0.5cm 0.5cm 1cm, clip=true, width=0.45\linewidth]{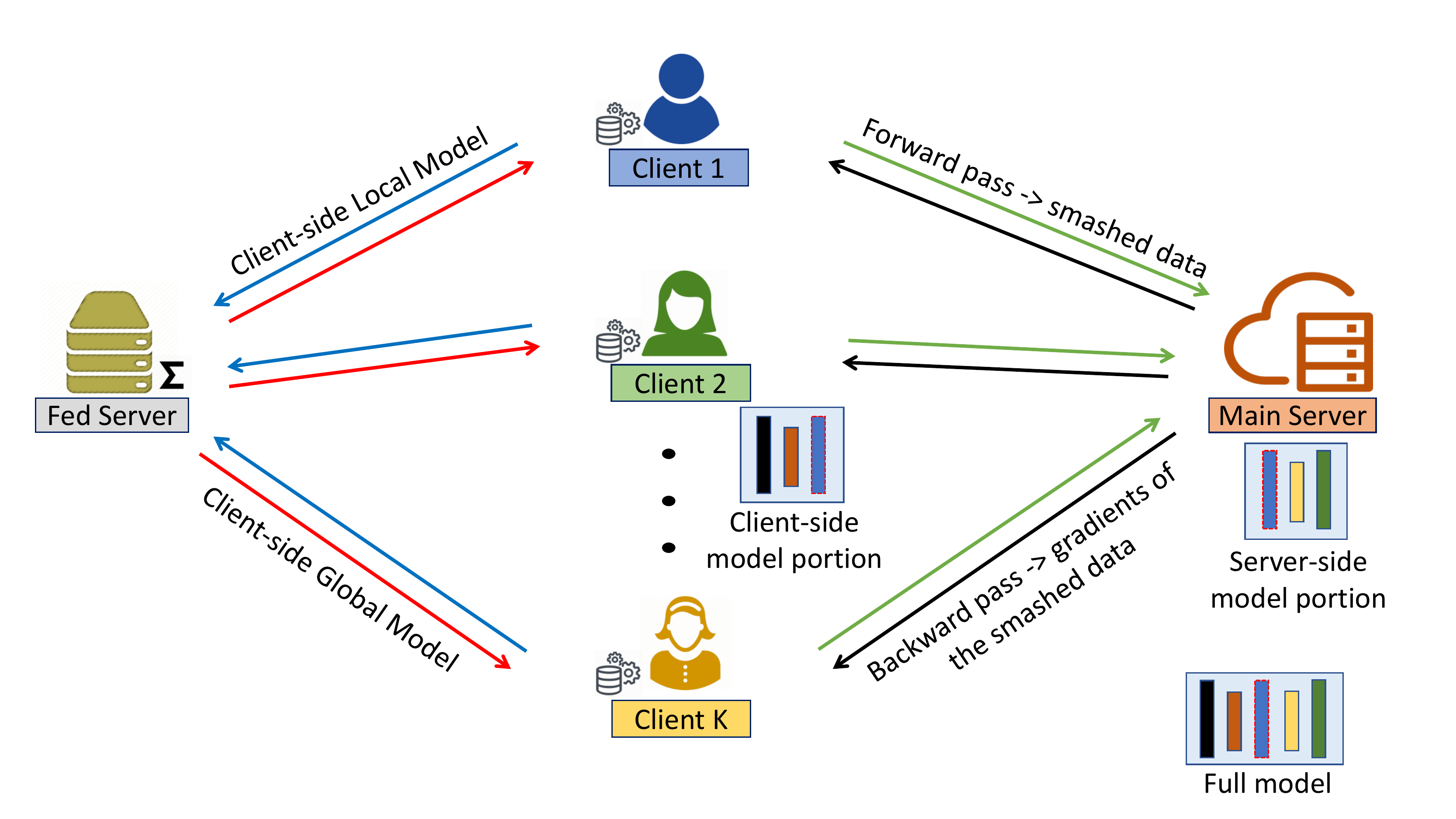}
			\caption{Splitfed learning architecture with $K$ clients and two servers; Fed server and Main server.}
			\label{fig:sec_splitfed_fig1}      
		\end{figure}
		
		SFL is a hybrid form of FL and SL. It performs client-side model training/testing in parallel, such as in FL, and trains/tests the full model by splitting it to the client-side and server-side for privacy and computation benefits such as in SL. Its architecture is divided into two main parts; the client-side part and the main-server part. In the client-side part, unlike SL, SFL introduces an extra worker, called the fed server, that is dedicated to performing the synchronization of the client-side model (see Fig.~\ref{fig:sec_splitfed_fig1}). The main server-side remains precisely the same as the server-side module in SL. 
		
		SFL operates in the following way:  

		\begin{algorithm} [!htb] 
			\small
			\SetNoFillComment
			\caption{\small Splitfed learning with label sharing~\cite{splitfed}. The notations are the same as in Algorithm~\ref{alg:split}.}
			\SetAlgoNoLine
			\SetKwProg{Fn}{EnsureMainServer executes at round $t\geq 0$:} {} {} \tcc{	\scriptsize Runs on Main Server}
			\Fn{}{ 
				\For{\textup{each client $ k\in S_t $ in parallel}} {
					$ (\mathbf{A}_{k,t}, \mathbf{Y}_{k}) \leftarrow$ ClientUpdate$(\mathbf{W}^{\textup{C}}_{k,t})$ \\
					Forward propagation with $ \mathbf{A}_{k,t} $ on $ \mathbf{W}^{\textup{S}}_t$, compute $ \hat{\mathbf{Y}}_{k} $  \\
					Loss calculation with $ \mathbf{Y}_{k}$ and  $\hat{\mathbf{Y}}_{k}$ \\
					Back-propagation calculate $\triangledown \ell_k (\mathbf{W}^{\textup{S}}_t; \mathbf{A}^{\textup{S}}_t)$\\
					Send $ d\mathbf{A}_{k,t} := \triangledown\ell_k (\mathbf{A}^{\textup{S}}_t; \mathbf{W}^{\textup{S}}_t) $ (i.e., gradient of the $ \mathbf{A}_{k,t} $) to client $ k $ for ClientBackprop($d\mathbf{A}_{k,t})$\\
				}
				Server-side model update: $ \mathbf{W}^{\textup{S}}_{t+1
				}  \leftarrow \mathbf{W}^{\textup{S}}_t - \eta\frac{n_k}{n} \sum_{i=1}^{K} \triangledown \ell_i (\mathbf{W}^{\textup{S}}_t;\mathbf{A}^{\textup{S}}_t)$\\}
			
			\vspace{5pt}
			\SetKwProg{Fn}{EnsureClientUpdate$  (\mathbf{W}^{\textup{C}}_{k,t}) $:}{}{} \tcc{	\scriptsize Runs on Client $ k $}	
			\Fn{}{
				Model updates $ \mathbf{W}^{\textup{C}}_{k,t}\leftarrow$ FedServer($ \mathbf{W}^{\textup{C}}_{t-1}$) \\
				Set $ \mathbf{A}_{k,t} $ = $ \phi $ \\
				\For{\textup{each local epoch $ e$ from $ 1 $ to $E$}}{
					\For{\textup{batch} $ b \in  \mathcal{B}$ }{
						Forward propagation on $ \mathbf{W}^{\textup{C}}_{k,b,t} $ \\
						Concatenate the activations of its final layer to $ \mathbf{A}_{k,t} $\\
						Concatenate respective true labels to $  \mathbf{Y}_{k}$\\ 
					}
				}
				Send $ \mathbf{A}_{k,t} $ and $\mathbf{Y}_{k}$ to the main server\\}	
			
			\vspace{5pt}
			\SetKwProg{Fn}{EnsureClientBackprop$  (d\mathbf{A}_{k,t}) $:} {} {}
			\tcc{	\scriptsize Runs on Client $ k $}
			\Fn{}{
				\For{\textup{batch} $ b \in  \mathcal{B}$}{
					Back-propagation, calculate gradients $ \triangledown  \ell_k (\mathbf{W}^{\textup{C}}_{k,b,t})$\\
					$\mathbf{W}^{\textup{C}}_{k,t}\leftarrow \mathbf{W}^{\textup{C}}_{k,t}-\eta \triangledown \ell_k (\mathbf{W}^{\textup{C}}_{k,b,t})$	
				}
				Send $ \mathbf{W}^{\textup{C}}_{k,t} $ to the Fed server}
			
			\vspace{5pt}
			\SetKwProg{Fn}{EnsureFedServer executes:}{}{}	 \tcc{	\scriptsize Runs on Fed Server}	
			\Fn{}{
				\For{\textup{each client $ k \in  S_t$ in parallel}}{
					$ \mathbf{W}^{\textup{C}}_{k,t} \leftarrow$ ClientBackprop($ d\mathbf{A}_{k,t} $) 
				}
				Client-side global model updates: $ \mathbf{W}^{\textup{C}}_{t+1}  \leftarrow \sum_{k=1}^{K} \frac{n_k}{n}  \mathbf{W}^{\textup{C}}_{k,t}$
				
				Send $\mathbf{W}^{\textup{C}}_{t+1} $  to all $K $ clients for ClientUpdate$  (\mathbf{W}^{\textup{C}}_{k,t}) $}
			\label{alg:splifed}
			
		\end{algorithm}

		%
		Firstly, the fed-server sends the global initial client-side model to all clients. Then, all clients (e.g., different hospitals or IoMTs with limited computing power) proceed with their forward propagation on their local data over the client-side model in parallel. Secondly, the smashed data are transmitted to the main server, which usually has enough computing power (e.g., cloud computing platform or research institution with a high-performance computing platform). Once the main server receives the smashed data from clients, it starts the forward propagation over the server-side model. The computations (forward and backpropagation) at the server-side associated with each client's smashed data can be done per client basis in parallel due to its high computing resource and independence of the operations related to clients. 
		The main server computes the gradients of the smashed data with respect to the loss function in its backpropagation and sends the gradients to the respective client. Thirdly, each client completes its backpropagation on their client-side models. The forward propagation and backpropagation between the clients and the server proceed for some rounds without the fed server. Afterward, the clients transmit the client-side networks' updates in the form of gradients to the fed server. Then, the fed server aggregates the updates and makes a global client-side model, which is sent back to all clients. This way, the client-side model synchronization happens in splitfed learning. Averaging is a simple method widely used for model aggregation in fed server; thus, the computation in it is not costly, making it suitable to operate within the local edge boundaries. There are two ways to do the server-side model synchronization: firstly, train the server-side model separately over the smashed data from each client, and later aggregate (e.g., a weighted average) all resulting server-side models to make the global server-side model -- this is known as splitfedv1 in~\cite{splitfed}-- and secondly, keep training the same server-side model over the smashed data from different clients -- this is known as splitfedv2 in~\cite{splitfed}. Splitfedv1 is illustrated in Algorithm~\ref{alg:splifed}, which is extracted from~\cite{splitfed}.

		
		\subsubsection{Key results in splitfed learning}
		There are three aspects that have been investigated in SFL: performance, training time latency, and privacy. Splitfed shows the same communication efficiency as of SL, with significantly less training time than SL~\cite{splitfed}. Empirical results considering ResNet18, AlexNet, and LeNet over HAM10000, FMNIST, CIFAR10, and MNIST datasets show that SFL (both splitfedv1 and splitfedv2) has comparative performance\footnote{Comparative performance refers to the case where two results are close to each other, and any result can be slightly higher or lesser than the other.} to SL and FL. Moreover, even for multiple clients ranging from 1 to 100 with ResNet18 over the HAM10000 dataset, the performance pattern is close to each other. However, this result is not present in general, specifically, for a higher number of clients. LeNet5 on FMNIST  with 100 users has slow convergence in splitfedv1 than others, and AlexNet on HAM10000 with 100 users has failed to converge in splitfedv2. For uniformly distributed HAM10000 and MNIST datasets with multiple clients over the ResNet18 and AlexNet, SFL is shown to reducing the training time by four to six times than SL. The same communication cost for both SL and SFL is observed~\cite{splitfed}. The privacy aspects of SFL is covered in Section~\ref{sec:splitfed_privacy}.
		
		\section{Data privacy and privacy-enhancing techniques}
		\label{sec:dataprivacy}
		
		FL, SL, and SFL provide certain privacy to the raw data, called default privacy, in their vanilla setting as the raw data are always within the control of data custodians, and no raw data sharing among the participants. This setup works perfectly well in an environment where the participating entities are semi-honest adversaries\footnote{A semi-honest adversary in a collaborative environment with multiple entities executes their assigned task as expected, but it can be curious about the information of the other entities.}. 
		However, if the capability of adversaries is increased more than their curious state, for example, membership inference attacks~\cite{shokri2017membership} and model memorization attacks~\cite{leino2020stolen}, then the default privacy is not sufficient, and other approaches are needed in addition. 
		%
		
		In literature, commonly used approaches are secure multiparty computation~\cite{mohassel2017secureml}, homomorphic encryption~\cite{rivest1978data}, and differential privacy (DP)~\cite{geyer2017differentially}.
		Secure multiparty computation
provides a privacy preservation model based on the zero-knowledge concept where multiple parties jointly compute a function on their inputs without letting their inputs to any other party~\cite{MPC2}.
		However, it is recommended under lower security requirements with a certain level of efficiency degradation, especially in communication cost~\cite{du2004privacy}. 
		Homomorphic encryption performs the computations of arbitrary functions on encrypted data without decryption. Thus an untrusted third-party entity can be involved in the computations while maintaining the privacy of input data/model~\cite{homomorphicsurvey,van2010fully}. However, due to the extensive computations and overhead, homomorphic encryption introduces a severe degradation of efficiency. Differential privacy involves the addition of calibrated noise to data or models so that the adversaries cannot extract private information while the expected utility of the data is maintained~\cite{dwork2008differential}. Due to the noise addition, differential privacy introduces a certain
		level of utility loss.

		By observing the side-by-side comparison, including computational efficiency and flexible scalability, of these approaches, differential privacy is considered the most favorable concept to improve data privacy~\cite{geyer2017differentially}. Furthermore, its properties such as immunity to post-processing, privacy guarantee, and composite differential privacy further signify its importance~\cite{diffprivacybook}. Differential privacy is formally defined in the following section. 

		\subsection{Differential privacy and its application}
		\label{sec:diffprivacy}
		Differential privacy (DP) is a privacy definition that constitutes a strong privacy guarantee~\cite{dwork2008differential,chamikara2019efficient}. A randomized algorithm, $\mathcal{K}$ provides differential privacy for $\delta \geq 0$ if for all adjacent datasets $D_1$ and $D_2$ (where $D_2$ differs from $D_1$ on at most one element) and all  $S \subseteq \textup{Range}(\mathcal{K})$,
		
		\begin{equation}
			\operatorname{Pr}\left[\mathcal{K}\left(D_{1}\right) \in S\right] \leq \exp (\varepsilon) \times \operatorname{Pr}\left[\mathcal{K}\left(D_{2}\right) \in S\right] + \delta,
		\end{equation}
		where $\varepsilon$ is called privacy budget that provides a measurement to the level of privacy leak form a certain function or algorithm $(\mathcal{K})$, which satisfies differential privacy~\cite{dwork2008differential,dwork2014algorithmic}. As the definition states, the value of $\varepsilon$ should be maintained at a lower level  (e.g., 0.1 to 9) to make sure that $\mathcal{K}$ does not leak an unacceptable level of information. $\delta$ provides a certain relaxation to the definition by providing a precalculated chance of failure. However, $\delta$ should be maintained at extremely low values (e.g., $1/(100*N)$, $N$: the number of instances in the database) to guarantee there is a meager chance (1\%) of privacy violation.  Applying noise over output results/queries to generate differentially private query/ML outputs is called global differential privacy. Applying noise on input data to generate differentially private datasets is called local differential privacy.

		
		
		\begin{figure}[tbh]
			\centering
			\includegraphics[scale=.15]{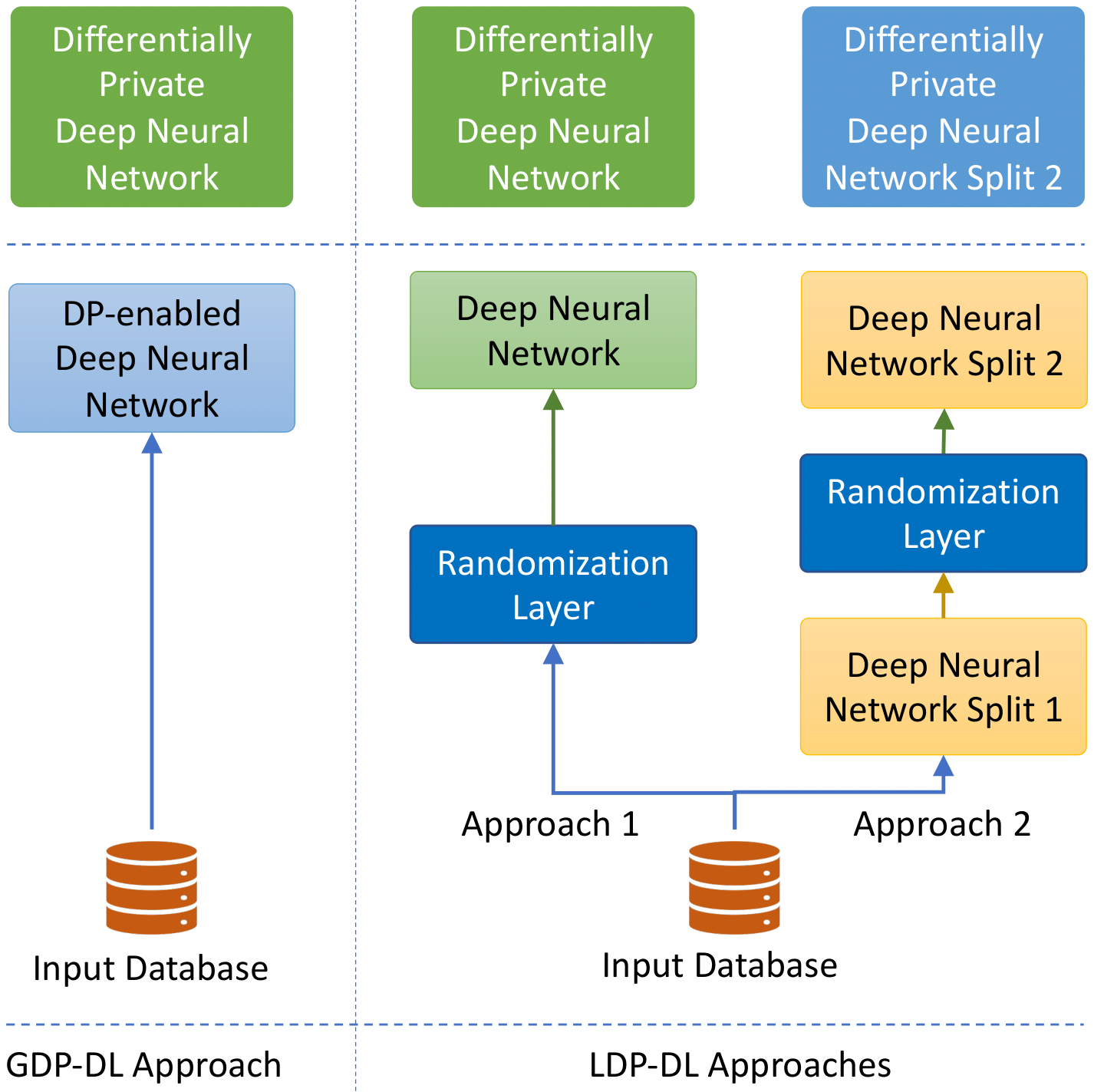}
			\caption{Different configurations of differentially private deep learning under global and local settings. DP: Differential Privacy/Differentially Private, DL: Deep Learning}
			\label{fig:1}       
		\end{figure}

		Due to the strong privacy guarantee, DP has been applied with deep learning~\cite{mohammed2011differentially} applications in DCML for areas such as healthcare~\cite{arachchige2020trustworthy}. Moreover, with DP, deep learning can guarantee a robust resistance to privacy attacks such as membership inference attacks and model memorization attacks~\cite{shokri2017membership,leino2020stolen}.  
		%
		%
		The differentially private solutions for deep learning can be categorized into two types: (1) approaches based on global differential privacy~\cite{abadi2016deep}, and (2) approaches based on local differential privacy~\cite{arachchige2019local}.
		Moreover, the existing architectural configurations of differentially private deep learning is depicted in Fig.~\ref{fig:1}. Global differential privacy applies noise based on the training algorithm. For example, it adds calibrated noise over the gradients of the model at each step of the stochastic gradient descent~\cite{abadi2016deep}. In contrast, local differential privacy applies noise to the data transferred between entities. For example, the noise can be added by adding an intermediate layer of randomization in between the convolutional layer and the fully connected layer of a convolutional neural network~\cite{arachchige2019local}. The global differential privacy has been the most popular approach for deep learning as the amount of noise added on the learning process is lower than the local differential privacy, which often adds overly conservative noise to maintain DP. In addition, the global differential privacy provides more flexibility in calibrating noise during the training process of a deep learning model. However, due to higher noise levels, the local differential privacy provides higher privacy levels than the global differential privacy~\cite{kairouz2014extremal}. The literature shows solutions under both types of differential privacy with high accuracy; however, local differential privacy solutions offer higher privacy guarantees. 

		\subsection{Differential Privacy in federated learning}
		\label{secfedpriv}

		Differential privacy is commonly integrated with FL, and it can be identified in two different forms: (1) adding differentially private noise to the parameter updates at the clients~\cite{wei2020federated}, (2) adding differentially private noise to the sum of all parameter updates at the server~\cite{geyer2017differentially}. In the first approach, calibrated noise is added to local weight (at the client-sides), whereas in the second approach, calibrated noise is added to global weight updates (at the server-side). Equation~\ref{eq:fl} represents the differentially private parameter update at the server model (for the second form above).  In this equation, $\Delta w_{k,t}$ is the client $k$'s parameter update at time instant $t$, $K$ is the number of clients, $S$ is the clipping threshold (sensitivity), $\mathcal{N}$ is the noise scaled to $S$. In the first form above, the noise addition mechanism follows a similar format. However, since the noise is added to each client model separately, $K$ becomes 1, where the parameter update considers only the current client model weights.
		\begin{equation}
			w_{t+1}=w_{t}+\frac{1}{K}\left(\sum_{k=0}^{K} \Delta w_{k,t} / \max \left(1, \frac{\left\|\Delta w_{k}\right\|_{2}}{S}\right)+\mathcal{N}\right)
			\label{eq:fl}
		\end{equation}

		\subsection{Privacy in split learning}
		\label{sec:splitprivacy}
		
		In contrast to FL, SL takes one step further to split the full model and its execution between clients and the server. Consequently, SL introduces an additional level of privacy to the full model, while training/testing, from the semi-honest clients and server~\cite{fedsurvey,Split_firstpaper}. This is because the main server has access only to the smashed data rather than the whole client-side model updates, and it is highly unlikely to invert all the client-side model parameters up to the raw data if the configuration of the client-side model portion has a fully connected layer with sufficiently large numbers of nodes~\cite{Split_firstpaper}. For other cases, there exist some possibilities, but this problem can be addressed by modifying the loss function at the client-side as in equation~\ref{eq:loss}~\cite{split_leakage2}. Due to the same reason as above, the client, who has access only to the gradients (of the smashed data) from the server-side, is not able to infer the server-side model portion.


		Besides, the SL configuration that trains a model without label sharing with the server (refer to Section \ref{sec:diffconfsplit}) enhances the inherent privacy by leaving the server clueless about classification results. However, the smashed data transferred from client to server still reveal a certain level of information about the underlying data. In this regard, differential privacy~\cite{ourpaper2,splitfed} and distance correlation techniques (see section~\ref{sec:inforleakage}) have been used in SL.  

		\subsection{Privacy in splitfed learning}
		\label{sec:splitfed_privacy}
		
		SFL aims to introduce a more efficient approach with a higher level of privacy by utilizing the advantages of both FL and SL~\cite{splitfed}. 
		%
		%
		%
		%
		The introduction of a local model federation server within the local bounds in the SFL architecture not only improves efficiency (e.g., training time) but also enhances privacy during the local model parameter synchronization~\cite{splitfed}. The enhanced privacy is obtained because the client has access only to the aggregated client-side model updates rather than untouched client-side model updates as in SL.
		To further limit privacy leaks, SFL investigates the use of differential privacy during local model training based on the differentially private deep learning approach developed in~\cite{abadi2016deep}, which we call ADPDL. In SFL, ADPDL for local model training is applied according to the following equation, 
		\begin{equation}
			\tilde{\mathbf{g}}_{k,t} \leftarrow \frac{1}{n_k} \sum_{i}\left(\overline{\mathbf{g}}_{k,t}\left(x_{i}\right)+\mathcal{N})\right),
			\label{dippeq3}
		\end{equation}
		which adds calibrated noise ($\mathcal{N}$) to the average gradient (where $\overline{\mathbf{g}}_{k,t}$ represents the $\ell_{2}$-norm clipped gradients) calculated in each step of the SGD algorithm. Next, the client side model parameters are updated according to $\mathbf{W}^{\textup{C}}_{k,t+1} \leftarrow \mathbf{W}^{\textup{C}}_{k,t}-\eta_{t} \tilde{\mathbf{g}}_{k,t}$ (where $\eta_{t}$ represent the learning rate) by taking a step back in the opposite direction.

		As a client is holding only a portion of the full model, the ADPDL will only have the full effect on the client-side model portion when the total budget is utilized. Hence, in the initial steps, the effect of noise over the activations and smashed data is minimal. To avoid any privacy leak due to this aspect, SFL adds a noise layer after the client-side model's cut layer. This layer adds calibrated noise over the smashed data in a utility-preserving manner based on Laplace mechanism~\cite{phan2017adaptive}. For this, the bounds ($ \text{max}\ \mathbf{A}_{k,i},\ \text{min}\ \mathbf{A}_{k,i}$) of smashed data, and a vector of intervals $\Delta I_i = \text{max}\ \mathbf{A}_{k,i}-\text{min}\ \mathbf{A}_{k,i}$  are calculated. Next, Laplacian noise with scale $\frac{\Delta I_i}{\varepsilon'}$ is applied to randomize the smashed data according to the following equation, 
		\begin{equation}
			\mathbf{A}^\textup{P}_{k,i}= \mathbf{A}_{k,i}+\operatorname{Lap}\left(\frac{\Delta I_i}{\varepsilon'}\right),
			\label{noiseq1}
		\end{equation}
		where $\epsilon'$ is the privacy budget used for the Laplacian noise.

		\section{Applications and implementation}
		\label{sec:applications}
		Applications of privacy-preserving machine learning approaches such as SL are widespread (e.g., finance and health) due to their inherent data privacy mechanisms. In this section, focusing on SL, firstly, the applications are presented, and then it provides the implementations from the programming perspective.
		\subsection{Applications of split learning:}
		
		\begin{figure}[H]
			\centering
			\includegraphics[trim=1.5cm 1cm 2cm 1cm, clip=true, width=0.5\linewidth]{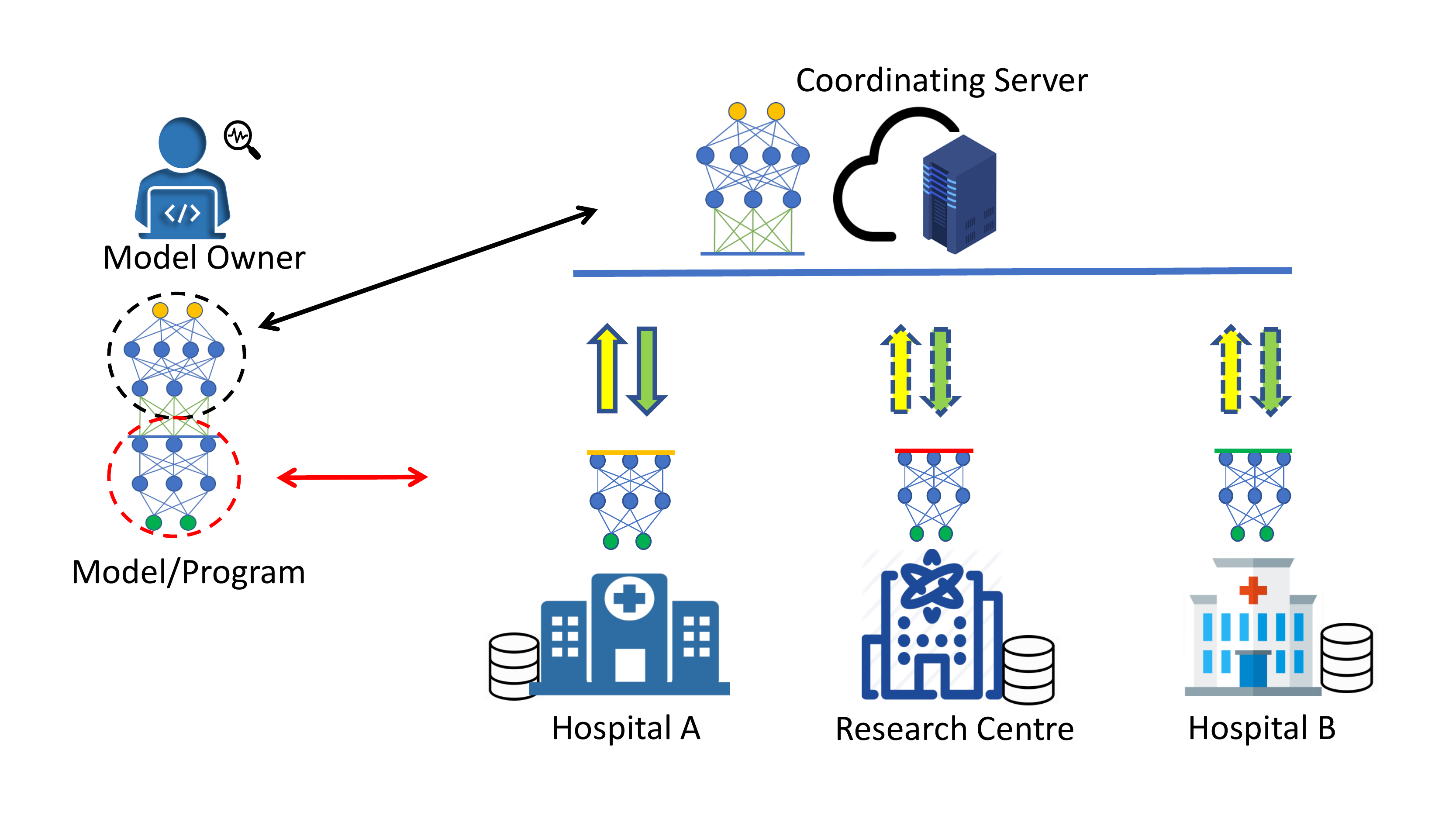}
			\caption{An application of split learning in a cross-siloed health environment.}
			\label{fig:split_app}      
		\end{figure}
		
		Health data analytics have been becoming an integral part of healthcare nowadays. The analytics are powered by (deep) machine learning/artificial intelligence models. However, to train these models, usually, a considerable amount of data are required. Though the health domain has sufficient data for ML training/testing, they are generally curated in silos due to privacy concerns. Thus, distributed machine learning approaches with privacy-preservation properties, such as SL, provide better alternatives for such domains. Fig.\ref{fig:split_app} illustrates a simple setup where a collaborative ML/AI training/testing is carried out among Hospital A, Research Center, and Hospital B without sharing their raw data with each other. This setup directly reflects the architectural configurations of SL. The primary benefits of SL over FL in this setting are model privacy and fewer client-side computations.    
		
		SL has been proposed for various settings with practical scenarios~\cite{split_differentconfiguration}. Techniques such as differential privacy and other noise integration methods are combined with SL to reducing the distance correlation of the smashed data with the raw input data for robust privacy. An improved SL with the KL divergence technique (section~\ref{sec:inforleakage}) is tested over colorectal histology images without any data augmentation~\cite{split_leakage1}. In a distributed setting with up to fifty clients, the proposed approach is used to analyze the diabetic retinopathy dataset over ReNet34 and chest X-ray dataset over DenseNet121. The results show that it provides better accuracy than non-collaborative techniques~\cite{poirot2019split}. In separate work, ECG data privacy is increased by integrating SL with the differential privacy measures and increasing the layers in the client-side model~\cite{ourpaper2}.
		
		SL is implemented for an edge-device machine learning configuration that is capable of effectively handling the internet of things (IoT) gateways. In a setup with five Raspberry pi kits as a gateway of IoTs, SL is evaluated over FL. SL shows significant results by reducing the training time -- 2.5 hours to run MobileNet on CIFAR10, whereas FL takes 8 hours-- and reduces the dissipated heat by the kit~\cite{ourpaper1}. This has practical importance as the kit can breakdown due to excessive heat due to the computations.
		
		SL is also successfully implemented in the wireless communication~\cite{Koda_2019}. Precisely, an SL based approach is proposed to integrate the image and radio frequency received signal, which is later used to predict the received power of millimeter-wave radio-frequency signals. This approach is a communication-efficient and privacy-preserving as it compresses the communication payload by adjusting the pooling size of the machine learning network architecture.  
		
		\subsection{Implementation of split learning}
		\label{sec:split_learning_implement}
\begin{figure}[!th]
			\centering
			\subfigure[]{
				\includegraphics[height=0.45\columnwidth]{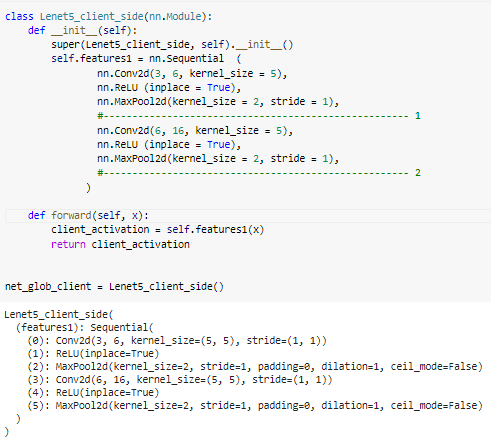}
			}
			\hskip-5pt
			\subfigure[]{
				\includegraphics[height=0.45\columnwidth]{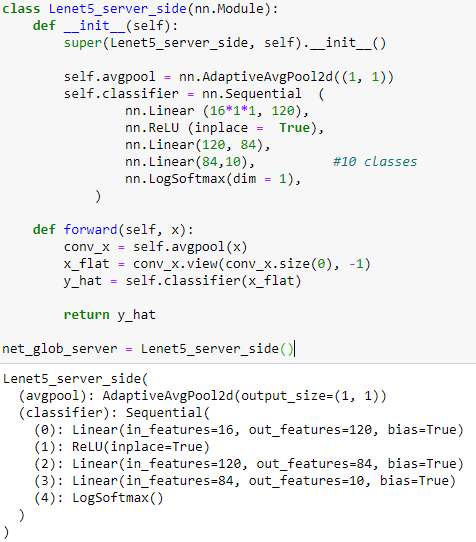}
			}
			
			\caption{A code snippet of an example of network architecture for the (a) client-side and (b) server-side portion of LeNet5, where the network split is done at the second layer after the max pool layer of the LeNet5 architecture.}
			\label{fig:client_server_architecture}
		\end{figure}

\begin{figure}[!hbt]
			\centering
			\subfigure[]{
				\includegraphics[height=0.12\columnwidth]{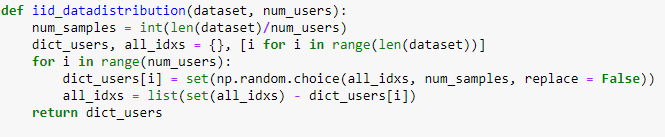}
			}
			\hskip-5pt
			\subfigure[]{
				\includegraphics[height=0.15\columnwidth]{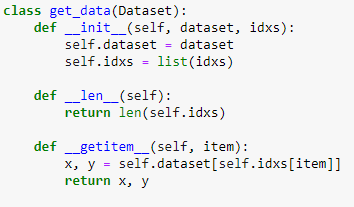}
			}
			
			\caption{A code snippet of a function and a class that is implemented for IID data distribution among clients; (a) returns a dictionary whose keys are the clients \emph{idx}, and lists of random indices of the samples as their values and (b) a class that extracts each data (i.e., x) and its label (i.e., y) in the whole Dataset, and its object is used as an input while creating a PyTorch data loader.}
			\label{fig:iid_datadistribution}
		\end{figure}

\begin{figure}[!hbt]
			\centering
			\includegraphics[height=0.4\columnwidth]{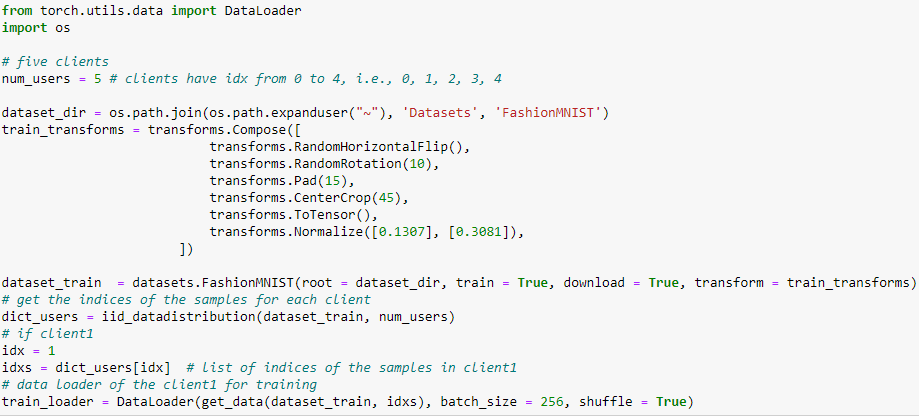}
			\caption{A code snippet depicting the implementation of the data loader in pytorch utilizing \texttt{iid\_datadistribution(dataset, numusers)} and \texttt{get\_data(Dataset)}.}
			\label{fig:dataloader}
		\end{figure}

\begin{figure}[!hbt]
			\centering
			\subfigure[]{
				\includegraphics[height=0.25\columnwidth]{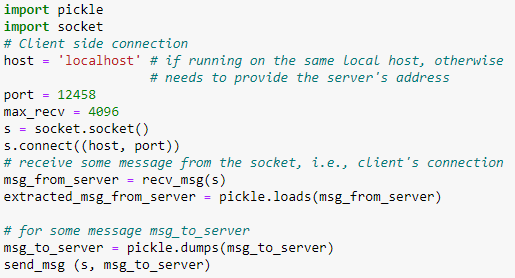}
			}
			\hskip-5pt
			\subfigure[]{
				\includegraphics[height=0.25\columnwidth]{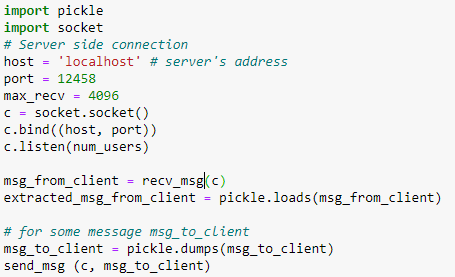}
			}
			
			\cprotect\caption{A code snippet of socket implementation in (a) client side, and (b) server side. \verb|recv_msg(connection)| and \verb|send_msg (connecton, msg)| are user-defined helper functions.}
			\label{fig:socket}
		\end{figure}

\begin{figure}[th]
			\centering
			\includegraphics[height=0.2\columnwidth]{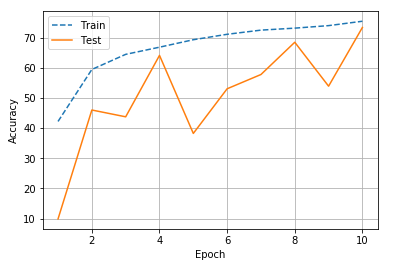}
			\caption{Accuracy curves of training and testing of LeNet5 on FMNIST dataset up to ten global epochs having one local epoch with the client each global epoch.}
			\label{fig:acc_curv}
		\end{figure}

		In this section, vanilla SL with LeNet5 on the FMNIST dataset is implemented for an illustration purpose. To this end, one client and one server are considered; however, the program can handle multiple clients. The programs for multiple clients is made by simply changing the device identity indicated by variable \emph{idx} in the same client program repeatedly for the respective client. The network architecture, i.e., LeNet5, is split to form the client-side network portion and the server-side network portion, as shown in Fig.~\ref{fig:client_server_architecture}. For multiple clients, the dataset is uniformly, identically, and independently distributed among them (the IID setting). One user-defined function \texttt{iid\_datadistribution(dataset, numusers)} and one class \texttt{get\_data(Dataset)} are implemented for this purpose, and their code snippets are provided in the Fig.~\ref{fig:iid_datadistribution}. The next step is to create a data loader, which is a combiner of both a dataset and a sampler, to iterate over the given dataset in PyTorch. This process is depicted in Fig.~\ref{fig:dataloader}.
		Each client's program and the server program run separately. Socket programming is used to simulate the communication between clients and the server. A Python socket is used along with some helper function for handling messages while the transmitting and receiving. Refer to Fig.~\ref{fig:socket} for a code snippet. All programs can be run on the same localhost or in different hosts. If different hosts, then the address of the host needs to be provided. For the initial setup, the server program runs first; then, the client programs are started sequentially.
		The complete program is available at GitHub\footnote{Some implementation by other sources: For extended SL and vertically partitioned SL: \url{https://github.com/nin-ed/Split-Learning},  for TensorFlow implementation: \url{https://colab.research.google.com/drive/1GG5HctuRoaQF1Yp6ko3WrJFtCDoQxT5_#scrollTo=esvT5OgzG6Fd}. Now available on PySyft.}, and the convergence curve for the accuracy while training and testing are provided in Fig~\ref{fig:acc_curv}. This result is obtained at the server-side after ten global epochs\footnote{One global epoch occurs if (forward and back) propagation is completed for all active clients' datasets for one cycle.}, where each global epoch has only one local epoch\footnote{In one local epoch of a client, one forward-propagation and its respective back-propagation are completed over entire local dataset of the client.}. Our complete implementation is available at~\cite{our_codesplit}.

\begin{figure}[t]
			\centering
			\subfigure[]{
				\includegraphics[height=0.12\columnwidth]{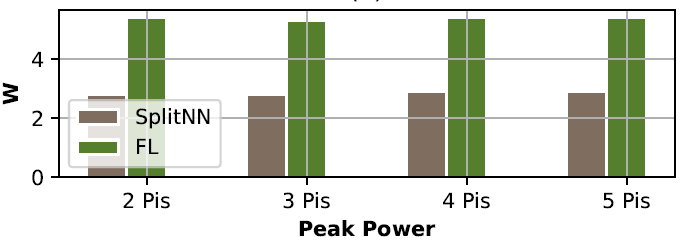}
			}
			\hskip-2pt
			\subfigure[]{
				\includegraphics[height=0.12\columnwidth]{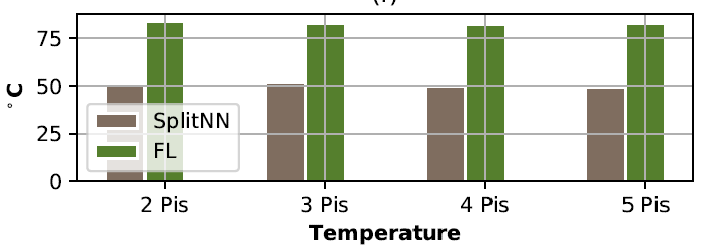}
			}
			
			\cprotect\caption{(a) Peak power and (b) temperature measurement~\cite{ourpaper1}. A plug-in power-meter is used to measure the power consumption in kWh unit. The temperature of the Raspberry Pi CPU is measured using a Python library \verb|CPUTemperature| and \verb|CPUTemperature()| function. The experiment is performed with 1D CNN (two CNN layers at the client-side, and one CNN and two fully-connected layers) model over the ECG dataset.}
			\label{fig:peakpower_and_temp}
		\end{figure}

		\paragraph{\textbf{Split learning implementation in Raspberry pi:}} 
		Raspberry pi kits are IoT gateways of low-end IoT devices. They support the corresponding operating systems (OS) to compute and process the ML/AI model on the client-side. 
		Raspberry Pi 3 model BV1.2 can run PyTorch version 1.0.0, OS Raspbian GNU/Linux 10, and python version 3.7.3 with no CUDA support. A useful manual to install Pytorch on Raspberry Pi is available at~\cite{install}. 
		In one of our works, a laptop having CPU i7-7700HQ, GPU GTX 1050, PyTorch version 1.0.0 installed, Windows 10, python version 3.6.8 installed, and CUDA version 10.1 was considered as the server. 
		For simplicity, a simple model architecture with 1D CNN on sequential time series data is trained on a setup with five Raspberry pi kits and a laptop. The kit and laptop were connected by 10Gbit/s dedicated LAN. Measurements of power and temperature per kit are depicted in Fig.~\ref{fig:peakpower_and_temp}. For more results and a complete code, refer to the paper~\cite{ourpaper1} and~\cite{code}, respectively.   
		
		\subsection{Implementation of splitfed learning with a code example}
		
		\begin{figure}[!tbh]
			\centering
			\subfigure[]{
				\includegraphics[height=0.2\columnwidth]{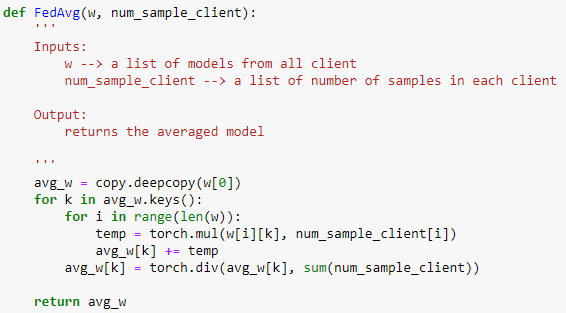}
			}
			\hskip-5pt
			\subfigure[]{
				\includegraphics[height=0.2\columnwidth]{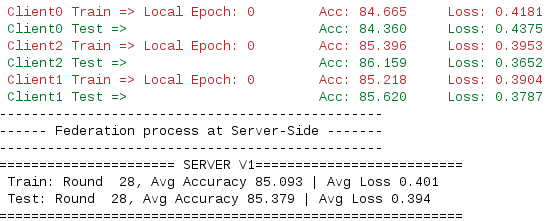}
			}
			
			\caption{(a) Code snippet of a user-defined function to compute weighted averaging of the models' weights received from multiple clients and collected in a list \texttt{w}, and (b) output snapshot while training at the main server at global epoch 28.}
			\label{fig:fed_avg}
		\end{figure}

		\begin{figure}[th]
			\centering
			\includegraphics[height=0.2\columnwidth]{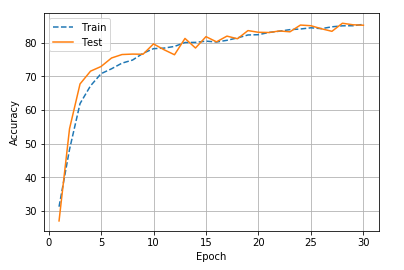}
			\caption{Average accuracy curves while training LeNet5 on FMNIST dataset up to 30 global epochs, each with one local epoch, over three clients.}
			\label{fig:acc_curv_splitfed}
		\end{figure}
		
		In this section,  vanilla SFL with LeNet5 on the FMNIST dataset is implemented. The setup has three clients and two servers (the main server and the fed server). There are three separate programs; one for the client, one for the fed server, and the other for the main server. The server program is capable of handling any number of clients. Clients are labeled by a unique identity indicated by positive integer starting from zero. The program for a client is obtained by simply assigning the variable \emph{idx} to the client's label in the program. The LeNet5 architecture is split as shown in Fig.~\ref{fig:client_server_architecture}. The dataset distribution part and the socket connection part are kept the same as in SL implementation (refer to Section~\ref{sec:split_learning_implement}). The server programs run first; then, the client program is started in the following sequence: fed server program, then main server program, after that all client programs. The complete program code is available at~\cite{our_codesplitfed}. The model aggregation at the fed server and the main server is done by applying weighted averaging on all locally trained model portions. The function that performs the aggregation is given as a code snippet in Fig.~\ref{fig:fed_avg}(a). A snapshot of the outputs at the server while training with respect to the global epoch is provided in Fig.~\ref{fig:fed_avg}(b) while Fig.~\ref{fig:acc_curv_splitfed} depicts the accuracy convergence during the training process. We use cross-entropy loss as our loss function. It measures the difference between the probability distributions of the ground truths and the outputs of the model.

		\section{Challenges and open problems}
		\label{sec:challenges}
		
		By keeping a focus on SL, this section presents and discusses the challenges and open problems in FL, SL, and SFL. This helps to shape future possible research avenues.
		
		\subsection{Challenges and open problems in federated learning}
		
		In FL, most of its vulnerabilities or privacy challenges come from untrusted entities. The entity can be a server or a client. Moreover, these entities can initiate various attacks, including model inversion attack~\cite{fredrikson2015model}, reconstruction attacks, membership-inference attacks~\cite{triastcyn2020federated} in FL environment. To overcome this vulnerability, literature shows the application of privacy and security-enhancing techniques such as DP, secure multi-party computation, and homomorphic encryption~\cite{geyer2017differentially,mohassel2017secureml,rivest1978data}. However, these are not enough in a general setting, as they suffer performance degradation like in DP, or high computation overhead like in homomorphic encryption. Thus, finding an optimal and practically feasible solution is still open.       
		%
		%
		Another prominent issue in FL is the need for constant communication with the central server. However, the distributed clients (most often) might have limited bandwidth and connectivity with the server. Hence, maintaining secure channels, maintaining a steady connection, and maintaining trust upon clients are extremely complex open challenges~\cite{reyzin2018turning}. FL is primarily based on the SGD optimization algorithm, which is used to train deep neural networks~\cite{zhao2018federated,fedsurvey}. To guarantee an unbiased estimate of the full gradients, the data should follow IID properties. However, in real-world scenarios, it is unlikely that the datasets follow IID properties, and real-world datasets often have non-IID properties. Consequently, with skewed non-IID data, FL can show poor performance~\cite{zhao2018federated}. Thus, how to perform FL under a highly skewed non-IID setting in a general setting is still an active field of research. However, the technique such as \emph{Federated Averaging with Spreadout (FedAwS)} are proposed to carry federated learning with only positive labels and embedding based classifiers~\cite{yu2020federated}.  
		
		\subsection{Challenges and open problems in split learning}    
		
		Due to the similarities in the distributed learning architectures, SL shares similar issues with FL. As the fundamental definition being the split of a network, SL's current configurations are limited to neural network-based model architectures~\cite{fedsurvey}. Along with the non-neural based architectures such as support vector machines, and even neural network-based networks such as recurrent neural networks (RNNs), it is yet to find an efficient way to split such networks to perform SL over them. However, for RNNs, split learning has been performed over the multiple clients, each containing consecutive RNN segments and a segment of sequential data~\cite{abedi2020fedsl}.  
		Similar to the cases of FL with non-IID data distribution, as mentioned in~\cite{fedsurvey}, SL is also susceptible to the data skewness. In terms of label distribution skew, where each client has samples all belonging to some class labels, SL also can show poor performance. Besides, SL does not learn under a one-class setup. 
		Compared to FL, SL may need more communications due to the additional step of local model synchronization (peer-to-peer or client-server). SL is also sensitive to data communication challenges. Similar to FL, maintaining secure channels, maintaining steady connections, and maintaining the trust upon clients are additional challenging problems in SL. Moreover, the unpredictability of a client's status (e.g., connected, dropped, sleeping, and terminated) play a significant role in the model convergence as the client model synchronization in SL is done through a passive sequential parameter update. 
		%
		
		The feasibility of any approach in a resource-constrained environment (e.g., smart city using IoTs) is driven by the communication efficiency of that approach. Herein, SL needs further attention. Compressing smashed data and reducing the number of communications necessary for the convergence in SL are two of the important research ventures. 
		Another main problem that is yet to solve fully is related to information leakage from the cut layer and countermeasures. Though there have been some works in this regard, it still needs further investigations to provide a potentially feasible solution that can limit the information leakage and maintain the utility of the model at the same time in general setup. 
		Client vulnerabilities (e.g., participation of vulnerable clients), server vulnerabilities (e.g. malicious servers), and communication channels issues (e.g., spoofing) are three of the architectural vulnerabilities of SL that need further investigations. Moreover, efficient incorporation of homomorphic encryption, secure multi-party computation, and DP to solve such vulnerabilities is still an open problem.
		

		
		
		
		
		

		\subsection{Challenges and open problems in splitfed learning}             
		
		SFL is devised by amalgamating SL and FL. However, the central concept of SFL is based on SL, whereas FL is used as an effective solution for the local model synchronization. Hence, apart from the local model synchronization issues in SL, all other challenges and open problems related to SL are also common to SFL.            
		
		
		
		\section{Conclusion}
		\label{sec:conclusion}
		This chapter presented an analytical picture of the advancement in distributed learning paradigms from federated learning (FL) to split learning (SL), specifically from SL's perspective. One of the fundamental features common to FL and SL is that they both keep the data within the control of data custodians/owners and do not require to see the raw data.
		In addition to this feature, SL provides an additional capability of enabling ML model training/testing in resource-constrained client-side environments by splitting the model and allowing the clients to perform computation only on a small portion of the ML model. Besides, full model privacy is achieved while training/testing is done by a curious server or clients in SL.   
		FL and SL both were shown to perform well over each other in different setups and configurations. For example, SL has faster convergence, however failing to converge if the data distribution is highly skewed. Whereas FL shows some resilience in this case, and it is possibly due to its model aggregation technique, which is weighted averaging. Thus, in general, SL, FL, and a hybrid approach like splitfed learning have their own importance. 
		As SL is in its initial development phase, there are multiple open problems related to several aspects, including communication efficiency, model convergence, device-based vulnerabilities, and information leakage. With research inputs over time, SL will become a matured privacy-preserving distributed collaborative machine learning approach.
		\bibliography{cas-refs}
		\bibliographystyle{plain}
		
	\end{document}